\documentclass[11pt]{article}

\newif\ifarxiv
\arxivtrue
\ifarxiv
  \usepackage[preprint]{acl}
\else
  \usepackage[review]{acl}
\fi
\input{acl_lineno_fix}

\usepackage{times}
\usepackage{latexsym}
\usepackage[T1]{fontenc}
\usepackage[utf8]{inputenc}
\usepackage{microtype}
\usepackage{inconsolata}

\usepackage{url}            %
\usepackage{booktabs}       %
\usepackage{amsfonts}       %
\usepackage{nicefrac}       %
\usepackage{xcolor}         %

\usepackage{graphicx}
\ifdefined\pdfinclusioncopyfonts\pdfinclusioncopyfonts=1\fi
\usepackage{subcaption}

\usepackage{appendix}

\usepackage{pifont}
\usepackage{amsmath}

\usepackage{tcolorbox}
\tcbuselibrary{skins,breakable}
\usepackage{tikz}
\usetikzlibrary{positioning,arrows.meta,shapes.misc,fit,calc}

\usepackage{enumitem}
\usepackage{wrapfig}  %

\usepackage{array}
\usepackage{tabularx}

\usepackage{lipsum} %
\newtcolorbox{promptbox}[2][]{
  enhanced,
  breakable,
  colback=gray!5,
  colframe=gray!60,
  boxrule=0.5pt,
  arc=4pt,
  outer arc=4pt,
  boxsep=2pt,
  left=6pt,
  right=6pt,
  top=4pt,
  bottom=4pt,
  listing only,
  listing options={basicstyle=\ttfamily, breaklines=true},
  fontupper=\ttfamily\scriptsize,
  fontlower=\ttfamily\scriptsize,
  title={#2},#1
}

\definecolor{lightpink}{RGB}{252, 232, 232}
\definecolor{lightblue}{RGB}{240, 245, 255}
\definecolor{lightgreen}{RGB}{232, 255, 224}

\definecolor{lightpurple}{RGB}{240, 233, 255}
\definecolor{lightorange}{RGB}{252, 229, 205}
\definecolor{lightyellow}{RGB}{255, 242, 204}
\definecolor{beige}{RGB}{228, 225, 214}
\definecolor{wine}{RGB}{234, 209, 220}

\let\origcolorbox\colorbox
\renewcommand{\colorbox}[2]{%
  {\setlength{\fboxsep}{0pt}%
     \origcolorbox{#1}{\hspace{3pt}#2\hspace{3pt}}}%
     }

\usepackage{color, colortbl}

\usepackage{multirow}

\newcolumntype{Y}{>{\centering\arraybackslash}X}

\definecolor{clrCtrl}{HTML}{8A5A00}    \definecolor{clrCtrlBg}{HTML}{FCEFC7}
\definecolor{clrLatent}{HTML}{5B2C8F}  \definecolor{clrLatentBg}{HTML}{EDE4F6}
\definecolor{clrExt}{HTML}{B34700}     \definecolor{clrExtBg}{HTML}{FBE4D2}
\definecolor{clrNeutral}{HTML}{455A64} \definecolor{clrNeutralBg}{HTML}{ECEFF1}
\definecolor{clrPass}{HTML}{2E7D32}    \definecolor{clrPassBg}{HTML}{E3F2E1}
\definecolor{clrFail}{HTML}{C62828}    \definecolor{clrFailBg}{HTML}{FBE3E3}
\colorlet{clrTopic}{clrCtrl}      \colorlet{clrTopicBg}{clrCtrlBg}      %
\colorlet{clrAgent}{clrCtrl}      \colorlet{clrAgentBg}{clrCtrlBg}      %
\colorlet{clrState}{clrLatent}    \colorlet{clrStateBg}{clrLatentBg}    %
\colorlet{clrPairing}{clrNeutral} \colorlet{clrPairingBg}{clrNeutralBg} %
\colorlet{clrConv}{clrExt}        \colorlet{clrConvBg}{clrExtBg}        %
\colorlet{clrOutcome}{clrExt}     \colorlet{clrOutcomeBg}{clrExtBg}     %

\newcommand{\hlctrl}[1]{\colorbox{clrCtrlBg}{#1}}
\newcommand{\hllatent}[1]{\colorbox{clrLatentBg}{#1}}
\newcommand{\hlext}[1]{\colorbox{clrExtBg}{#1}}

\newcommand{\hlpass}[1]{\colorbox{clrPassBg}{\textcolor{clrPass}{#1}}}
\newcommand{\hlfail}[1]{\colorbox{clrFailBg}{\textcolor{clrFail}{#1}}}

\newcommand{\stg}[2]{%
\colorbox{#1Bg}{\strut\textcolor{#1}{\textbf{\footnotesize\sffamily#2}}}}

\newcommand{\passbadge}{\colorbox{clrPassBg}{\textcolor{clrPass}{\small\bfseries\sffamily\ding{51}~PASS}}}
\newcommand{\failbadge}{\colorbox{clrFailBg}{\textcolor{clrFail}{\small\bfseries\sffamily\ding{55}~FAIL}}}

\newcommand{\cpass}{\cellcolor{clrPassBg}\textcolor{clrPass}{\ding{51}}}
\newcommand{\cfail}{\cellcolor{clrFailBg}\textcolor{clrFail}{\ding{55}}}
\newcommand{\cpassinline}{\hlpass{\ding{51}}}
\newcommand{\cfailinline}{\hlfail{\ding{55}}}
\newcommand{\passrate}[1]{\ifnum#1>4\textcolor{clrPass}{\textbf{#1/9}}\else\ifnum#1=0\textcolor{clrFail}{\textbf{#1/9}}\else\textcolor{clrFail}{#1/9}\fi\fi}

\newcommand\vr[1]{}

\newcommand{\cpdelete}[1]{}
\newcommand{\my}[1]{}

\title{Are LLM Agents Behaviorally Coherent? A Latent Profiling Approach for Social Simulation}

\newcommand{\projectlinks}{}
\ifarxiv
\renewcommand{\projectlinks}{\\[0.5em]
  \raisebox{-0.4ex}{\includegraphics[height=1em]{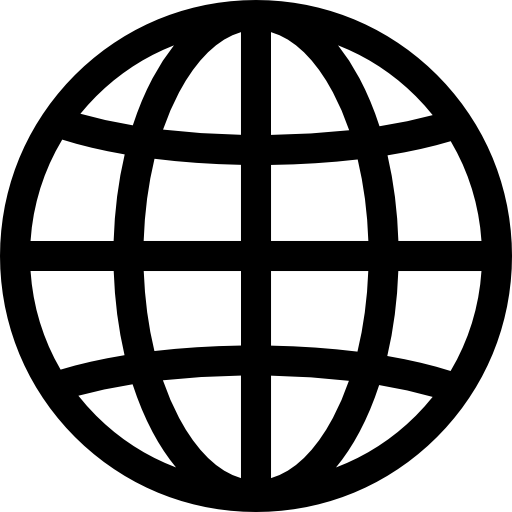}}\hspace{0.3em}\href{https://minnesotanlp.github.io/latent_profile/}{\texttt{Website}}
  \hspace{0.2cm}
  \raisebox{-0.4ex}{\includegraphics[height=1em]{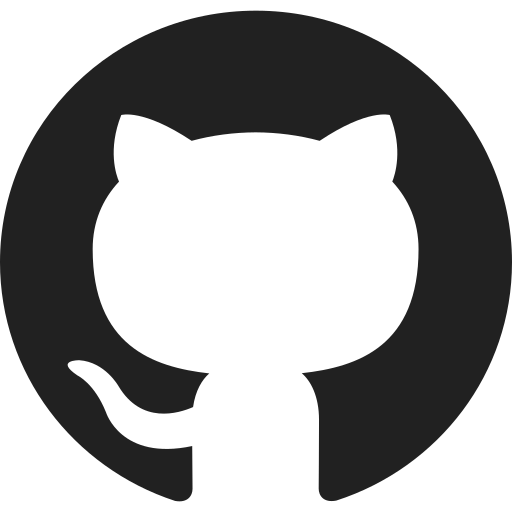}}\hspace{0.3em}\href{https://github.com/minnesotanlp/latent_profile}{\texttt{Code}}}
\fi

\author{
James Mooney\textsuperscript{1}$^{*}$, Josef Woldense\textsuperscript{2}$^{*}$,
Zheng Robert Jia\textsuperscript{1}, Shirley Anugrah Hayati\textsuperscript{1}, \\
\textbf{My Ha Nguyen\textsuperscript{3}, Vipul Raheja\textsuperscript{4}, Dongyeop Kang\textsuperscript{1}} \\
\textsuperscript{1}Department of Computer Science and Engineering, University of Minnesota \\
\textsuperscript{2}Department of African American \& African Studies, University of Minnesota \\
\textsuperscript{3}Department of Sociology, University of Chicago,
\textsuperscript{4}Grammarly \\
\texttt{\{moone174,jwoldens,dongyeop\}@umn.edu}\projectlinks
}

\newcommand{\fw}{Latent Profiling}

\ifarxiv
\fi

\begin{document}

\maketitle
\ifarxiv
{\renewcommand{\thefootnote}{\fnsymbol{footnote}}\footnotetext[1]{Equal contribution.}}
\fi

\begin{abstract}
    LLM agents are increasingly proposed as substitutes for human participants in social research, usually on the strength of survey responses that match those of the humans they represent. We argue that such output-level alignment is not enough. Drawing on the behavioral sciences, we propose \fw{}, a framework for testing \textit{behavioral coherence}: whether an agent's conversational behavior follows, as human behavioral models predict, from its own \textit{latent profile} elicited by short surveys. As a case study, we test six hypotheses about preference, openness and agreement across nine open models. Agents reproduce surface-level trends but systematically fail in-depth tests: they are not behaviorally coherent, even when their individual responses look human. \fw{} thus offers a theory-grounded instrument for a behavioral science \emph{of} LLMs.
\end{abstract}

\section{Introduction}

LLMs have demonstrated remarkable progress in recent years, prompting researchers and practitioners alike to ask not whether these systems can pass the Turing test \citep{jones2025large}, but whether they can convincingly adopt full-fledged human personas \citep{hu-collier-2024-quantifying, park2023generative}.
Early findings suggest they can. For example, \citet{park2024} find that when agents are constructed using rich qualitative interview data, they exhibit attitudes and behaviors that closely mirror those of their human counterparts. Such results have inspired what we term the \textit{substitution thesis}: if agents can emulate humans, they may serve as substitutes for real participants in human-centered research. As substitutes, agents can be examined for individual traits or can be deployed to simulate human societies at scale. Should this prove viable, the potential upsides for social research would be tremendous:
companies might test new products on virtual customers \citep{SimUser, ExploreCustomer} , and social scientists could explore complex phenomena like war \cite{huawar}, governance \cite{piatti2024cooperate}, or cultural evolution \cite{perez2024cultural} with fewer ethical and logistical constraints.

\begin{figure}[t]
\centering
\begin{tikzpicture}[x=1cm,y=1cm,font=\footnotesize\sffamily,
  ev/.style={circle,line width=0.6pt,inner sep=0pt,minimum size=17pt,font=\small\sffamily},
  ctrl/.style={ev,draw=clrCtrl!80,text=clrCtrl},
  lat/.style={ev,draw=clrLatent,fill=clrLatentBg,text=clrLatent,line width=1.1pt,minimum size=21pt,font=\small\sffamily\bfseries},
  ext/.style={ev,draw=clrExt!80,minimum size=10pt},
  outc/.style={ev,draw=clrExt,text=clrExt,line width=0.8pt,minimum size=19pt},
  val/.style={font=\footnotesize\sffamily,text=black!65,align=center},
  grp/.style={font=\footnotesize\sffamily,align=center},
  flow/.style={-stealth,black!35,line width=0.5pt},
]
\node[ctrl] (c)  at (0.35,0)  {C};
\node[lat]  (p)  at (1.65,0) {P};
\node[lat]  (o)  at (2.85,0) {O};
\node[ext]  (t1) at (3.9,0) {};
\node[ext]  (t2) at (4.5,0)  {};
\node[ext]  (t3) at (5.1,0) {};
\node[outc] (a)  at (6.45,0) {A};
\foreach \u/\w in {c/p,p/o,o/t1,t1/t2,t2/t3,t3/a} \draw[flow] (\u) -- (\w);
\node[val] at (0.35,-0.58) {prompt};
\node[val] at (1.65,-0.58) {$+2$/$-2$};
\node[val] at (2.85,-0.58) {low};
\node[val] at (4.5,-0.58)  {turns};
\node[val] at (6.45,-0.58) {\textbf{agree}};
\node[grp,text=clrCtrl]   at (0.35,-1.05) {control};
\node[grp,text=clrLatent] at (2.25,-1.05) {\textbf{latent profile}};
\node[grp,text=clrExt]    at (5.2,-1.05) {conversation};
\draw[-stealth,clrLatent,line width=0.9pt] (2.25,0.42) .. controls (3.0,1.35) and (5.6,1.35) .. (a.north);
\node[font=\footnotesize\sffamily,text=clrLatent,fill=white,inner sep=1pt] at (4.3,1.13) {predicts \textbf{disagree}};
\node[font=\normalsize\bfseries,text=red!70!black] at (7.0,0.36) {\ding{55}};
\end{tikzpicture}

\caption{\textbf{Detecting behavioral inconsistency through latent profiles.} Each event looks plausible on its own: a \hlctrl{control} prompt ($C$), a conversation, and an agreement outcome ($A$). Only the agents' \hllatent{latent profile} from short surveys (preference $P$, openness $O$) reveals what a human behavioral model predicts (opposed, low-openness agents should disagree), exposing the observed agreement as inconsistent.}
\label{fig:motivation}
\end{figure}

\begin{figure*}[t]
    \centering
\begin{tikzpicture}[x=1cm,y=1cm,font=\small\sffamily,
  every node/.append style={execute at begin node={\hyphenpenalty=10000\exhyphenpenalty=10000}},
  panel/.style={rounded corners=4pt,line width=0.9pt},
  head/.style={font=\normalsize\sffamily,text=black,anchor=north west,inner sep=0pt},
  sub/.style={font=\footnotesize\sffamily,text=black!60,anchor=north west,inner sep=0pt},
  card/.style={rounded corners=2.5pt,draw=black!20,fill=white,line width=0.5pt,inner sep=3pt,align=left,font=\small\sffamily},
  av/.style={circle,inner sep=0pt,minimum size=15pt,font=\scriptsize\sffamily\bfseries,text=white},
  bub/.style={rounded corners=5pt,fill=white,draw=clrExt!50,line width=0.6pt,inner sep=3pt,font=\small\sffamily,align=left},
  big/.style={-stealth,line width=2pt,black!30},
]
\def\H{5.0}
\fill[panel,clrCtrlBg] (0,0) rectangle (3.6,\H);
\draw[panel,clrCtrl!60] (0,0) rectangle (3.6,\H);
\node[head] at (0.15,\H-0.15) {1\enspace Controls};
\node[sub] at (0.15,\H-0.62) {set in the prompt};
\node[av,fill=clrNeutral] at (0.45,3.55) {A1};
\node[card,anchor=west] at (0.72,3.55) {woman, 60s, rural};
\node[av,fill=black!50] at (0.45,2.85) {A2};
\node[card,anchor=west] at (0.72,2.85) {man, 30s, urban};
\node[card,anchor=south west,text width=3.1cm] at (0.12,0.18)
  {\textbf{Topic}\enspace Fall/Spring\\[2pt]
   \textbf{Bias}\enspace firm stance};

\fill[panel,clrLatentBg] (4.0,0) rectangle (8.05,\H);
\draw[panel,clrLatent,line width=1.9pt] (4.0,0) rectangle (8.05,\H);
\node[head] at (4.15,\H-0.15) {2\enspace Latent profile};
\node[sub] at (4.15,\H-0.62) {self-reported in a survey};
\node[fill=clrLatent,text=white,font=\scriptsize\sffamily\bfseries,rounded corners=2pt,inner sep=2pt,anchor=north east] at (7.9,\H-0.13) {KEY};
\node[font=\normalsize\sffamily\bfseries,text=clrLatent,anchor=west] at (4.15,3.3) {$P$};
\draw[clrLatent!35,line width=1.4pt] (4.75,3.3) -- (7.55,3.3);
\foreach \i in {0,...,4}{\fill[clrLatent!50] ({4.75+\i*0.7},3.3) circle (2.4pt);}
\node[av,fill=clrNeutral] at (4.75,3.7) {A1};
\node[av,fill=black!50] at (7.55,3.7) {A2};
\node[font=\footnotesize\sffamily,text=black!55,anchor=north] at (4.75,3.15) {Fall};
\node[font=\footnotesize\sffamily,text=black!55,anchor=north east] at (7.75,3.15) {Spring};
\node[font=\normalsize\sffamily\bfseries,text=clrLatent,anchor=west] at (4.15,2.0) {$O$};
\draw[clrLatent!35,line width=1.4pt] (4.75,2.0) -- (7.55,2.0);
\foreach \i in {0,...,4}{\fill[clrLatent!50] ({4.75+\i*0.7},2.0) circle (2.4pt);}
\node[av,fill=clrNeutral] at (4.75,2.4) {A1};
\node[av,fill=black!50] at (5.3,2.4) {A2};
\node[font=\footnotesize\sffamily,text=black!55,anchor=north] at (4.75,1.85) {closed};
\node[font=\footnotesize\sffamily,text=black!55,anchor=north east] at (7.75,1.85) {open};
\node[card,anchor=south,text width=3.6cm,align=center,draw=clrLatent!60] at (6.025,0.18)
  {opposed $+$ closed $\Rightarrow$\\ expect \textbf{disagreement}};

\fill[panel,clrExtBg] (8.45,0) rectangle (12.1,\H);
\draw[panel,clrExt!60] (8.45,0) rectangle (12.1,\H);
\node[head] at (8.6,\H-0.15) {3\enspace Conversation};
\node[sub] at (8.6,\H-0.62) {observed behavior};
\node[av,fill=clrNeutral] at (8.85,3.6) {A1};
\node[bub,anchor=west] at (9.12,3.6) {``Fall is the best.''};
\node[av,fill=black!50] at (11.75,2.85) {A2};
\node[bub,anchor=east] at (11.48,2.85) {``Spring, for sure.''};
\node[av,fill=clrNeutral] at (8.85,2.1) {A1};
\node[bub,anchor=west] at (9.12,2.1) {``Fair point, I agree!''};
\node[card,anchor=south,text width=3.2cm,align=center,draw=clrExt!60] at (10.275,0.18)
  {agreement per turn\\ $A{=}(2,3,4,5)$\;\textcolor{clrExt}{$\nearrow$}};

\fill[panel,black!3] (12.5,0) rectangle (15.95,\H);
\draw[panel,black!25] (12.5,0) rectangle (15.95,\H);
\node[head] at (12.65,\H-0.15) {4\enspace Coherence test};
\node[sub] at (12.65,\H-0.62) {does behavior match?};
\node[font=\small\sffamily,anchor=west] at (12.65,3.6) {expected};
\fill[clrLatent!60,rounded corners=1pt] (12.7,3.0) rectangle (13.2,3.25);
\node[font=\small\sffamily,anchor=west,text=clrLatent] at (13.25,3.13) {low $A$};
\node[font=\small\sffamily,anchor=west] at (12.65,2.6) {observed};
\fill[clrExt,rounded corners=1pt] (12.7,2.1) rectangle (14.85,2.35);
\node[font=\small\sffamily,anchor=west,text=clrExt] at (14.9,2.23) {high};
\node[rounded corners=3pt,fill=clrFailBg,draw=clrFail!60,text=clrFail,inner sep=4pt,
      font=\normalsize\sffamily\bfseries,anchor=south] at (14.25,0.25) {\ding{55}\ incoherent};

\draw[big] (3.62,2.5) -- (3.98,2.5);
\draw[big] (8.07,2.5) -- (8.43,2.5);
\draw[big] (12.12,2.5) -- (12.48,2.5);
\draw[-{Stealth[length=9pt,width=9pt]},clrLatent,line width=2.2pt,rounded corners=8pt]
  (6.025,\H+0.05) -- (6.025,\H+0.4) -- (14.2,\H+0.4) -- (14.2,\H+0.1);
\node[anchor=south,inner sep=2pt,font=\small\sffamily\bfseries,text=clrLatent] at (10.1,\H+0.42)
  {the latent profile defines what behavior should look like};
\end{tikzpicture}
    \caption{\textbf{Overview of \fw{}, our framework for testing behavioral coherence}, shown with an illustrative example from our case study. (1) \hlctrl{Controls} set each agent's persona, the topic, and a stance instruction. (2) A short survey elicits each agent's \hllatent{latent profile}, its preference $P$ and openness $O$, which together with a human behavioral model define what its behaviour \emph{should} look like. (3) Agents converse, and their \hlext{external behaviour} is scored as per-turn agreement $A$. (4) Coherence tests check observed $A$ against the profile-based expectation. Each component can be swapped out for other settings.}
    \label{fig:high_level_pipeline}
\end{figure*}

Still, the gap between technological promise and practical utility remains large.
\citet{park2024} achieve high persona fidelity only with two-hour interviews; with less background information, persona mimicry drifts \citep{li2024measuring, abdulhai2025consistently, sandhan-etal-2026-persona}, and simulated populations match aggregate human opinions while showing far less variance \citep{bisbee2024synthetic} and ``flattened'' identity groups \citep{Wang2025_identity}.

\paragraph{From output alignment to behavioral coherence.}
Almost all of these evaluations ask the same question: \emph{do an agent's outputs look like those of the humans it represents?} We argue this is not enough. A human participant is useful to social science because their behavior across a study \emph{hangs together}: what they say in a survey predicts, through well-established behavioral models, how they act in a later interaction. We therefore ask a different question: \emph{is an agent's behavior consistent with its own revealed internal state?} This is a question behavioral science asks of people, and we argue it should be asked of LLM agents too, as part of a \emph{behavioral science of LLMs}. We call this property \emph{behavioral coherence}; throughout the paper, \emph{coherence} names the property and an \emph{inconsistency} is an observed violation of it. An agent can pass the first test and fail the second: every stage of a simulated study may look plausible on its own while the stages contradict each other (Figure~\ref{fig:motivation}).

\paragraph{Contributions.}
\begin{itemize}[leftmargin=*,itemsep=2pt,topsep=2pt]
  \item \textbf{A new evaluation target: behavioral coherence.} We move the question from output-level alignment with humans to \emph{internal} coherence, where an agent's conversational behavior must follow from its own self-reported state as human behavioral models predict.
  \item \textbf{The \fw{} framework} (Figures~\ref{fig:high_level_pipeline} and~\ref{fig:design_frame}). \fw{} separates what is measured about an agent into three streams: \hlctrl{controls} set in the prompt, a \hllatent{latent profile} elicited by short surveys of each agent in isolation, and \hlext{external interactions} observed in multi-agent conversation. It then tests the streams \emph{jointly} against hypotheses from human behavioral models. We designed the framework together with social and behavioral scientists, so that each test operationalizes an established model of attitudes, persuasion or group discussion rather than an ad hoc expectation. The framework is model- and domain-agnostic: other traits, interaction formats and behavioral models can be plugged in.
  \item \textbf{A case study across nine open models.} We instantiate \fw{} with six hypotheses about agreement, grounded in topic preference and openness, and apply it to the Gemma-3, Llama-3.x and Qwen3 families (0.6B--12B).
\end{itemize}

\paragraph{Key findings from the case study.}
\fw{} separates what agents reproduce from what they do not (Table~\ref{fig:results}; numbers link to the findings in Section~\ref{sec:results}).
(1) \textbf{Agents look human in aggregate} (\hyperref[find:1]{\#1}, \hyperref[find:5]{\#5}): agreement falls steadily as preferences diverge (3.76 at gap 0 to 1.90 at gap 4), and more open pairs agree more.
(2) \textbf{But they drift toward agreement regardless of who they are} (\hyperref[find:2]{\#2}, \hyperref[find:6]{\#6}): telling agents to hold a firm stance raises agreement even between opposed partners ($+0.20$), and among maximally opposed pairs the \emph{least} open agree the most.
(3) \textbf{The same latent profile does not produce the same behavior} (\hyperref[find:3]{\#3}, \hyperref[find:4]{\#4}): pairs who share a dislike agree far less than pairs who share a liking (2.6 vs.\ 4.8 at zero gap), and the topic alone shifts agreement by up to 0.71 points at identical preferences.
(4) \textbf{No model escapes}: across all nine models, each in-depth test passes for at most one model, while surface tests pass for most (Table~\ref{tab:results}).

These failures are invisible to output-level evaluation and appear only once conversational behavior is conditioned on each agent's latent profile. \fw{} is deliberately simple, yet it already exposes concrete weaknesses of LLM agents as behavioral subjects, which makes it a practical starting point for a behavioral science of LLMs. To support social and behavioral scientists in applying it to their own constructs and populations, we will publicly release our code, prompts, data and the full \fw{} pipeline.

\begin{figure*}[t]
    \centering
    \input{tex/fig_pipeline}
    \caption{\textbf{Data-collection pipeline of the framework.} \hlctrl{Controls}: (1) we choose topics $T$ of varying contentiousness $C$ and (2) build agent profiles from demographics $D$ and a topic-bias instruction $B$. \hllatent{Latent profile}: (3) each agent is surveyed in isolation for its preference $P$ on the topic and its openness $O$ to persuasion. \hlext{External interaction}: (4) agents $(U_m,U_n)$ are paired to discuss the topic, and (5) an LLM judge scores their agreement $A$ after every turn. The dashed arrow marks the core of the framework: the observed outcome is tested for coherence against the agents' latent profiles. Stage numbers match the pipeline-stage row of Table~\ref{fig:results}.}
    \label{fig:design_frame}
\end{figure*}

\section{Related Work}

\textbf{LLMs as human substitutes.}
LLMs show context-sensitive reasoning~\citep{wei2022chain, kojima2022large} and are increasingly used not only as tools but as experimental subjects~\citep{mou2024individual, park2023generative}. In multi-agent simulations they exhibit socially emergent behaviors~\citep{wang2025user} and have been used to model conformity and information cocoons~\citep{anthis2025llm}, war~\citep{huawar}, market competition~\citep{zhaoCompeteAI2024}, and group deliberation~\citep{yang2024confidence}; their survey responses can also align closely with human data~\citep{anthis2025llm}. Yet behavioral plausibility can mask instability: simulated responses lack human-like variance and demographic nuance and are sensitive to prompts and temporal drift~\citep{bisbee2024synthetic, petrov2024limited, takata2024spontaneous}. Cognitive evaluations point the same way: LLMs give inconsistent judgments on narratives~\citep{delangis2025temporal}, pair strong working memory with weak executive control~\citep{delangis2026executive}, and often miss characters acting against their stated traits~\citep{delangis2026mary}.

\textbf{Coherence of personality and preference.}
Fluent conversation does not imply continuity: LLMs lose identity- and preference-linked information across turns~\citep{castillo2024beyond}, struggle to act on preferences that are never stated explicitly~\citep{tsaknakis2026latent}, and topic-relevance benchmarks~\citep{fan2024topic} ignore how traits shape engagement. Persona-based dialogue models~\citep{zhang2018personalizing, rashkin2019towards} offer stylized variation without evolving, trait-informed reasoning, while generative agents with memory~\citep{park2023generative} and trust-aware simulations~\citep{xie2024can} do not test how personality shapes agreement. Persona conditioning improves coherence and broadens the spread of psychological traits in generated dialogue~\citep{wu2025personas, ge2024scaling}. Since Big Five traits such as openness are stable over time~\citep{cobb2012stability} and tied to cognitive flexibility~\citep{oshio2018resilience}, we ask whether agents' conversational behavior is consistent with such traits as measured from the agents themselves.

\section{A Framework for Probing Behavioral Coherence}
\label{sec:method}

To measure behavioral coherence, we design a novel experimental framework that probes how latent profiles of LLMs evolve through prompting and interaction. Rather than aiming to exhaustively characterize all possible behaviors, this framework serves as a prototype for systematically evaluating a specific class of behavioral dynamics. In particular, it captures how model-internal dispositions shift and manifest across conversational sequences.

\subsection{Experimental Framework}

Our framework has five sequential stages (Figure \ref{fig:design_frame}): (1) select a topic, (2) generate agents, (3) elicit their latent states, (4) pair them for dialogue, and (5) score agreement (Appendices \ref{sec:frame}, \ref{sec:agent-demo}). 

\paragraph{\textbf{Topic Selection:}} Construct a set of topics, where each topic $T$  is associated with a contentiousness level $C \in \{1, 2, 3\}$, with 1 being the least contentious and 3 being the most contentious . The set contains nine topics in total, with exactly three topics assigned to each contentiousness level. Further descriptions of the topics may be found in Table \ref{tab:topics}.
  
\paragraph{\textbf{Generate Agents:}} For each topic, construct agents with demographic profiles $D_i$ defined by age, gender, urbanicity, location, and education (see Appendix \ref{sec:agent-demo} for more details on specific prompt construction). We further modify the agent prompt to include their bias towards the topic at hand, $B \in \{1, 2, 3\}$, with 1 being the least biased towards an opinion on a topic, and 3 being the most biased. Demographic region is limited to the United States and systematically varied across 5 age groups, 2 genders, 4 regions, 4 urbanicity levels, and 6 education levels.
  
\paragraph{\textbf{Internal State:}} Identify each agent’s preference $P_i$ and openness $O_i$. Both these values are captured by posing a set of questions to agent $i$ and the responses are used to create a number indicating an agent's preference for a topic and their openness - both of which are represented as integers between 1 and 5. See Section \ref{sec:frame} for more details how these values are computed.

\paragraph{\textbf{Pairing Agents:}} In order to properly test a given set of behavioral models, we must ensure that we are sampling from a large space of agent pairings during conversation. To this end, we must provide a process mechanism for pairing agents in the \textit{External Interactions} portion of our framework.

As each agent $i$ has a unique preference, openness and bias associated with it, the \textit{controlled} profile of an agent may be given by $(P_i, O_i, B_i)$. Let $U := \{(P_j, O_j, B_j)\} \,\, \forall j \in \{1, .., N\}$ be the set of all agents. Then, for all possible pairs $(U_m, U_n)$, we sample agents $q, r$ such that $(P_q, O_q, B_q) = U_m$ and $(P_r, O_r, B_r) = U_n$. This means that we are mechanically preparing all possible combinations of preference, openness and bias to interact with one another - enabling higher fidelity when mapping to our behavioral models of interest.

\paragraph{\textbf{Conversation Outcome:}} For each step $K$ of a conversation, we use LLM-as-judge to score agreement $A \in \{1, 2, 3, 4, 5\}$ (1 = complete disagreement, 5 = complete agreement). For analysis, we retain only the final agreement score. To calibrate judgments, we provide five annotated sample conversations—one for each score.

\begin{table*}[t]
\centering
\newcommand{\famchip}[2]{\tikz[baseline=(n.base)]\node[fill=#1Bg,text=#1,rounded corners=2pt,inner xsep=2pt,inner ysep=1.4pt,font=\scriptsize\sffamily\bfseries](n){#2};}
\newcommand{\vdot}[1]{\tikz[baseline=-0.62ex]\fill[#1](0,0)circle(2.5pt);}
\newcommand{\nodot}{\tikz[baseline=-0.62ex]\fill[black!18](0,0)circle(0.9pt);}
\newcommand{\stagelbl}[1]{{\footnotesize\sffamily\color{black!45}#1}}
\newcommand{\fdesc}[1]{\hspace{0.5em}{\footnotesize\color{black!60}#1}}
\newcommand{\depth}[1]{{\footnotesize\sffamily #1}}
\small
\setlength{\tabcolsep}{2pt}
\renewcommand{\arraystretch}{1.0}
\begin{tabularx}{\linewidth}{@{}r@{\hspace{3pt}}>{\raggedright\arraybackslash}Xccccclc@{}}
\toprule
& & \multicolumn{2}{c}{\famchip{clrCtrl}{Controls}} & \multicolumn{2}{c}{\famchip{clrLatent}{Latent states}} & \famchip{clrExt}{External} & & \\
& \textbf{Behavioral-model test}
  & \textcolor{clrCtrl}{$C$} & \textcolor{clrCtrl}{$B$}
  & \textcolor{clrLatent}{$P$} & \textcolor{clrLatent}{$O$}
  & \textcolor{clrExt}{$A$} & \textbf{Depth} & \textbf{Verdict} \\[-2pt]
& \raisebox{2pt}{\stagelbl{pipeline stage $\rightarrow$}} & \stagelbl{[1]} & \stagelbl{[2]} & \stagelbl{[3]} & \stagelbl{[3]} & \stagelbl{[5]} & & \\
\midrule
\multicolumn{9}{@{}l}{\small\itshape On preferences (\S\ref{sec:ref-pref})}\\
\#1 & {Preference gap decreases agreement}\fdesc{similar $\Rightarrow$ agree more}
    & \nodot & \nodot & \vdot{clrLatent} & \nodot & \vdot{clrExt} & \depth{Surface} & \passbadge \\
\#2 & {Bias instruction asymmetry}\fdesc{bias does not polarize}
    & \nodot & \vdot{clrCtrl} & \vdot{clrLatent} & \nodot & \vdot{clrExt} & \depth{In-depth} & \failbadge \\
\#3 & {Shared sentiment, divergent destinations}\fdesc{sign matters}
    & \nodot & \nodot & \vdot{clrLatent} & \nodot & \vdot{clrExt} & \depth{In-depth} & \failbadge \\
\#4 & {Contentiousness at shared preference}\fdesc{topic shifts agreement}
    & \vdot{clrCtrl} & \nodot & \vdot{clrLatent} & \nodot & \vdot{clrExt} & \depth{In-depth} & \failbadge \\
    \midrule
\multicolumn{9}{@{}l}{\small\itshape On openness (\S\ref{sec:exp-open})}\\
\#5 & {Openness increases agreement}\fdesc{open $\Rightarrow$ agree more}
    & \nodot & \nodot & \nodot & \vdot{clrLatent} & \vdot{clrExt} & \depth{Surface} & \passbadge \\
\#6 & {Low openness + large gap $\Rightarrow$ lowest agreement}\fdesc{they over-agree}
    & \nodot & \nodot & \vdot{clrLatent} & \vdot{clrLatent} & \vdot{clrExt} & \depth{In-depth} & \failbadge \\
\midrule
\multicolumn{9}{c}{\hlctrl{Controls} $+$ \hllatent{Latent states} $+$ \hlext{External interactions} $\Longrightarrow$ \textit{coherence disintegrates}}\\
\bottomrule
\end{tabularx}
\caption{Main findings, aligned with our pipeline (Figure~\ref{fig:design_frame}). Each row is one human behavioral model; filled dots mark the artifacts it combines: \hlctrl{control variables} ($C$, $B$), \hllatent{latent states} ($P$, $O$) and the \hlext{external interaction} outcome ($A$). Verdicts are for \texttt{Gemma-3-12b-it}; see Table~\ref{tab:results} for other models. \textit{Surface}-level tests \hlpass{\ding{51}~pass}, whereas \textit{In-depth} tests that require latent states to hold under finer-grained conditions \hlfail{\ding{55}~fail}.}
\label{fig:results}
\end{table*}

\subsection{Specific Experiments}
We examine 9 topics with different controversy levels to explore the effect of contentiousness on both individual agent response as well as on agreement during interaction with other agents. Among these topics (in descending order of contentiousness) are opinions on "Taxes", "Remote Work" and "Coca Cola vs. Pepsi". More details on specific topics and agent demographics can be found in Appendix \ref{sec:frame} (in particular, Tables \ref{tab:topics} and \ref{tab:demographics} show the set of topics and agent demographics we consider).

Using these topics and demographics, we conduct two sets of experiments: (1) qualitative experiments (\S4.1 and \S4.2) as a proof-of-concept on LLMs' internal coherence on preferences and openness, and (2) robustness experiments (\S4.3) to provide statistical comparisons across various LLM model sizes. For both experiment sets, we use \texttt{Qwen3-32B} ~\citep{qwen3} as the judge. For all the qualitative visualizations below, we use \texttt{Gemma-3-12b-it}  \citep{gemma3} as the agent model. For the purposes of robustness, we examine the results of our approach across different sizes of models for the Gemma3 \citep{gemma3}, Llama3.x \citep{llama3}, and Qwen3 \citep{qwen3} model families.
\section{Experimental Results}
\label{sec:results}

In our experiments, we aim to uncover the discrepancy between the \textit{appearance} of behavioral coherence and its breakdown under closer examination with more complex tests. We organize our findings around \textbf{preference} and \textbf{openness} as both have well-established behavioral models of discussion and persuasion \citep{huckfeldt1995citizens, petty1986communication}. We further break our results down by whether the test is "Surface-level" or "In-depth". This is an implicit dimension, but typically relies upon looking at our data in a more granular fashion - moving from simpler statistical tests to those which are more complex. To reiterate, we are using the \texttt{Gemma-3-12b-it} models for all visualizations that follow. Table~\ref{fig:results} generally demonstrates our results across the six behavioral models discussed in detail below.
For more details on the precise nature of statistical tests discussed in this section, refer to Appendix \ref{app:statistical_tests}. We also provide a series of qualitative examples showing the agent profiles and conversation outputs for each of the findings in Appendix \ref{sec:qual-failures}.

\begin{figure*}[t!]
\centering

\begin{minipage}[t]{0.485\textwidth}
    \centering
    \includegraphics[width=\linewidth]{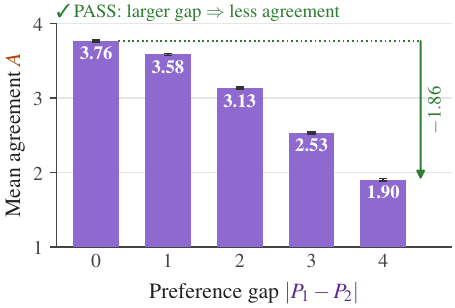}
    \caption{When two agents have a larger gap in their preference for a topic (e.g.\ favorite soda, views on taxes), they agree less in conversation: mean agreement drops by 1.86 points from gap 0 to gap 4.}
    \label{fig:pref_gap_agree}
\end{minipage}
\hfill
\begin{minipage}[t]{0.485\textwidth}
    \centering
    \includegraphics[width=\linewidth]{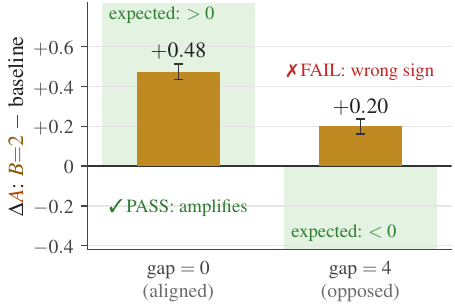}
    \caption{High bias ($B{=}2$) raises agreement at a preference gap of 0 as expected ($+0.48$), but also raises it at a gap of 4 ($+0.20$), where it should lower it. Shading marks the expected sign.}
    \label{fig:bias_asymmetry}
\end{minipage}

\end{figure*}

\subsection{On Preferences}
\label{sec:ref-pref}

\phantomsection\label{find:1}\paragraph{Finding \#1: Preference Gap Decreases Agreement}
We start with the the most basic coherence test: Agents with shared preferences should agree more than those with opposing preferences, the similarity--agreement link documented in political discussion networks \citep{huckfeldt1995citizens, mutz2006hearing}. We measure this using the \textit{preference gap} that is computed for each conversation pair. The latter is defined as the absolute difference between the agents’ preference scores. A gap of four indicates maximal divergence whereas a gap of zero indicates identical views.

Figure \ref{fig:pref_gap_agree} shows that the results are consistent with our expectation: pairs with aligned preferences (gap$=$0) achieve the highest agreement, while pairs with greater gaps yield progressively lower scores. A Pearson correlation between preference gap and agreement score confirms a significant negative relationship ($r < 0$, $p < 0.01$): agreement falls as the gap widens. 

\phantomsection\label{find:2}\paragraph{Finding \#2: Bias Instruction Asymmetry}
In our framework, recall that the bias term is used to influence the  relative strength of a belief around a given topic. In our case, the strongest bias is indicated by the bias parameter $B_i = 2$, which directs agents to adopt a strong stance either for or against the topic. A firm stance should amplify both directions: social judgment theory predicts assimilation of nearby positions and contrast with distant ones \citep{sherif1961social}, and committed partisans polarize further when confronted with opposing views \citep{lord1979biased}.

We evaluate this by comparing conversations where both agents receive the bias instruction ($B_1 = B_2 = 2$) against all other configurations, which serve as the baseline, at two extremes of the preference gap. As illustrated in Figure \ref{fig:bias_asymmetry}, effects are interpreted relative to this baseline where positive values indicate higher agreement than the baseline, while negative values indicate lower agreement. The expected pattern is directional -- at gap$=0$, the bar should shift upward (above baseline), whereas at the maximum gap of four, it should shift downward (negative below baseline). At gap$=0$, where agents share identical preferences, this expectation is borne out: bias amplifies agreement, with a one-sided Mann–Whitney U test confirming that $B=2$ pairs achieve significantly higher agreement than the baseline ($p < 0.01$, Bonferroni-corrected for two comparisons).

At the maximum gap of four, where agents hold maximally divergent preferences, the expectation is the opposite. The bar should move in the negative direction to indicate reduced agreement relative to the baseline. However, as shown in Figure \ref{fig:bias_asymmetry}, this pattern does not materialize. Rather than scoring below the baseline, agreement is actually higher, revealing a clear asymmetry in how bias instructions act across preference gaps.

\begin{figure}[!t]
\centering
\begin{tcolorbox}[enhanced, colframe=clrLatent, colback=white, boxrule=0.5pt, arc=2pt,
  left=4pt, right=4pt, top=2pt, bottom=2pt, fontupper=\footnotesize,
  colbacktitle=clrLatentBg, coltitle=clrLatent, fonttitle=\footnotesize\sffamily\bfseries,
  title={Latent profile: $P_1{=}P_2$ (gap $0$) $\Rightarrow$ human model expects high agreement in both}]
\textbf{(a) Shared negative, $P{=}(1,1)$} \hfill \textit{Spring vs.\ Fall}\\[1pt]
\textit{A1:} ``Nah, Fall's the best.''\quad
\textit{A2:} ``Absolutely not. Spring's just muddy and full of sneezes.''\\[1pt]
\textcolor{clrExt}{Agreement} $A{=}(2,2,2,2)$ \hfill \textcolor{red!70!black}{\ding{55}\ low, despite zero gap}\\[-3pt]
\textcolor{black!20}{\rule{\linewidth}{0.4pt}}\\[-1pt]
\textbf{(b) Shared positive, $P{=}(5,5)$} \hfill \textit{Taxes}\\[1pt]
\textit{A1:} ``Absolutely, it's how we fund things like schools and infrastructure.''\quad
\textit{A2:} ``Oh, absolutely. It's how we pay for things like schools and roads.''\\[1pt]
\textcolor{clrExt}{Agreement} $A{=}(5,5,5,5)$ \hfill \textcolor{clrPass}{\ding{51}\ high, as expected}
\end{tcolorbox}

\caption{\textbf{Qualitative example (Finding \#3).} Both pairs share the same preference (gap $0$), so their \hllatent{latent profiles} predict the same high agreement. Read in isolation, each conversation is fluent and on-persona; only by conditioning on the latent profile does the asymmetry surface: shared aversion yields persistent low agreement while shared support yields full agreement. Examples for all six findings are in Appendix~\ref{sec:qual-failures}.}
\label{fig:qual_example}

\end{figure}

\begin{figure*}[t]
    \centering
    \includegraphics[width=\textwidth]{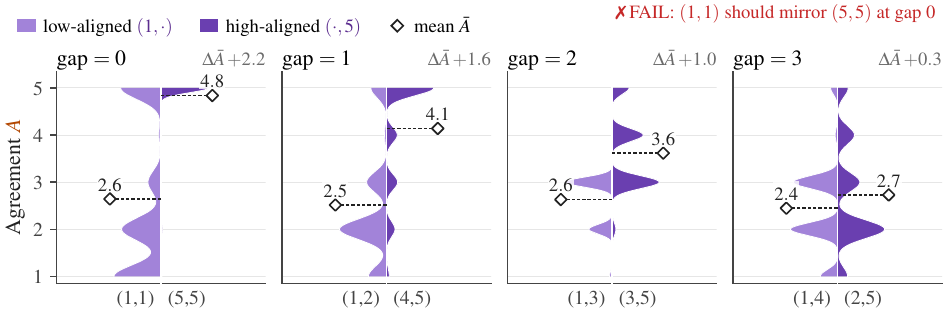}
    \caption{Agreement distributions of low-aligned $(1,\cdot)$ pairs (light) and high-aligned $(\cdot,5)$ pairs (dark) at each preference gap; $\diamond$ marks the mean. The distributions do not match: even at gap 0, $(1,1)$ pairs agree far less than $(5,5)$ pairs, contrary to what behavioral models suggest.}
    \label{fig:pref_agree_less}
\end{figure*}

\phantomsection\label{find:3}\paragraph{Finding \#3: Shared Sentiment, Divergent Destinations}
Consider two pairs of agents: one pair strongly favors a topic (preferences 5,5), while the other strongly dislikes it (preferences 1,1). In principle, both pairs should exhibit high agreement -- one through shared enthusiasm, the other through shared aversion. Balance theory treats a shared dislike as just as unifying as a shared liking \citep{abelson1968theories}, so the two pairs should agree comparably. Figure~\ref{fig:qual_example} shows a representative pair of conversations: both are fluent and on-persona, and the inconsistency only becomes visible once they are read against the agents' latent profiles.

Contrary to expectation, Figure \ref{fig:pref_agree_less} reveals a striking directional bias in agreement outcomes. For nearly all gaps, high-aligned pairs $(\cdot, 5)$ outperform the corresponding low-aligned pairs $(1, \cdot)$, showing that agreement depends on the direction of alignment. Strikingly, despite having a gap of 0 and theoretically expected to achieve the highest agreement scores, pair (1,1) consistently scores below other pairs with larger gaps. These patterns are formally confirmed by one-sided Mann–Whitney U tests with Bonferroni correction, which show that low-aligned pairs rarely achieve significantly higher agreement than their high-aligned counterparts across the gap spectrum, highlighting a pronounced inconsistency in how shared sentiment translates into agreement.

\begin{figure*}[t!]
\centering

\begin{minipage}[t]{0.485\textwidth}
    \centering
    \includegraphics[width=\linewidth]{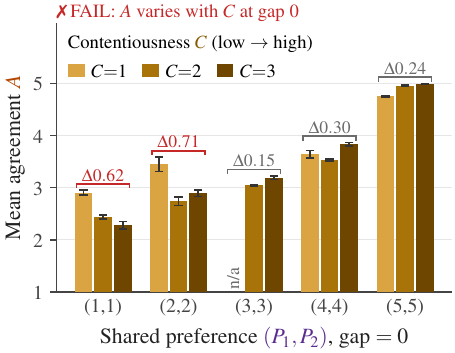}
    \caption{Agreement varies with topic contentiousness $C$ even for identical preferences (gap $=0$), especially for shared negative preferences. Brackets give the spread across $C$ within each pair, which should be $\approx 0$.}
    \label{fig:content_share_preference}
\end{minipage}
\hfill
\begin{minipage}[t]{0.485\textwidth}
    \centering
    \includegraphics[width=\linewidth]{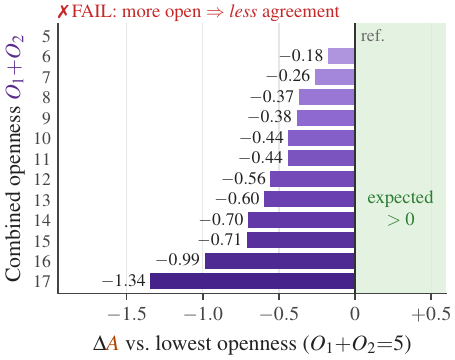}
    \caption{With maximal preference gaps, lower openness yields \emph{higher} agreement: every more-open pairing agrees less than the least-open one ($O_1{+}O_2{=}5$), the opposite of the expected sign (shaded).}
    \label{fig:openness_low_agreement}

\end{minipage}

\end{figure*}

\phantomsection\label{find:4}\paragraph{Finding \#4: Contentiousness at Shared Preference}
In principle, when both agents hold identical views on a topic, the inherent contentiousness of that topic should not independently affect agreement. 
The similarity--attraction paradigm locates agreement in attitude similarity itself rather than in the topic \citep{byrne1971attraction}.

Translated into the current study, we fix the preference gap at zero (i.e., both agents share the same preference) to isolate the role of topic characteristics. Our objective is to determine whether changes in topic contentiousness ($C = 1, 2, 3$) lead to systematic differences in agreement. Under the prior literature, the null hypothesis makes a clear prediction: when preferences are aligned, the distribution of agreement should remain effectively unchanged across levels of contentiousness -- any observed differences should be negligible and attributable to random variation rather than the topic itself.

Contrary to this expectation, Figure \ref{fig:content_share_preference} shows that, even with the preference gap held constant, agreement scores vary meaningfully across levels of contentiousness. This is formally corroborated using the Kruskal–Wallis test, which assesses whether agreement distributions differ across groups.

\subsection{On Openness}
\label{sec:exp-open}
\phantomsection\label{find:5}\paragraph{Finding \#5: Openness Increases Agreement}
At the aggregate level, the relationship between openness and agreement aligns with expectations. As the combined openness of a conversation pair increases, average agreement also rises, as the elaboration likelihood model predicts for more receptive audiences \citep{petty1986communication, brinol2023openness}.
We formalize this with a Pearson correlation between combined openness (the sum of both agents' openness scores) and agreement. The test confirms a significant positive relationship ($r > 0$, $p < 0.01$), indicating that openness reliably increases agreement in aggregate. However, as with preferences, this surface-level finding masks deeper inconsistencies.

\phantomsection\label{find:6}\paragraph{Finding \#6: Low Openness and Large Preference Gap Produces Lowest Agreement}

\begin{table}[t]
\centering
\footnotesize
\setlength{\tabcolsep}{2.4pt}
\renewcommand{\arraystretch}{1.06}
\begin{tabular}{@{}l cccc c >{\centering\arraybackslash}p{3.0em}>{\centering\arraybackslash}p{3.0em}@{}}
\toprule
& \multicolumn{4}{c}{\stg{clrState}{Preference $P$}} & & \multicolumn{2}{c}{\stg{clrState}{Openness $O$}} \\
\cmidrule(lr){2-5}\cmidrule(l){7-8}
& {\scriptsize Surf.} & \multicolumn{3}{c}{\scriptsize In-depth} & & {\scriptsize Surf.} & {\scriptsize In-depth} \\
Model & T1 & T2 & T3 & T4 & & T5 & T6 \\
\midrule
Qwen3-0.6B   & \cfail & \cfail & \cfail & \cfail & & \cfail & \cfail \\
Qwen3-4B     & \cfail & \cfail & \cfail & \cfail & & \cpass & \cfail \\
Qwen3-8B     & \cpass & \cfail & \cfail & \cfail & & \cpass & \cpass \\
\midrule
Llama-3.2-1B & \cfail & \cfail & \cfail & \cfail & & \cpass & \cfail \\
Llama-3.2-3B & \cpass & \cfail & \cfail & \cfail & & \cpass & \cfail \\
Llama-3.1-8B & \cpass & \cpass & \cfail & \cfail & & \cpass & \cfail \\
\midrule
Gemma-3-1B   & \cfail & \cfail & \cfail & \cfail & & \cfail & \cfail \\
Gemma-3-4B   & \cpass & \cfail & \cfail & \cfail & & \cpass & \cfail \\
Gemma-3-12B  & \cpass & \cfail & \cfail & \cfail & & \cpass & \cfail \\
\midrule
\textit{Pass rate} & \passrate{5} & \passrate{1} & \passrate{0} & \passrate{0} & & \passrate{7} & \passrate{1} \\
\bottomrule
\end{tabular}
\vspace{1mm}
\caption{Results of the six behavioral-model tests across LLMs (T$k$ = Finding~\#$k$; \cpassinline~pass, \cfailinline~fail).}
\label{tab:results}

\end{table}

If openness functions as intended, then the combination of low openness and large preference divergence should produce the lowest possible agreement -- motivated-reasoning accounts predict that closed, strongly opposed partners discount each other's arguments the most \citep{lord1979biased, kunda1990motivated}. We test this by isolating conversations with maximal preference divergence (pairs 1,5) and comparing the agreement distribution of the lowest-openness pairing against all other openness pairings within this subset.

\label{sec:formalized}

Specifically, we identify the openness pairing with the smallest combined openness score and test whether its agreement distribution falls significantly below each alternative pairing. We apply one-sided Mann--Whitney U tests with Bonferroni correction across all comparisons, requiring that the low-openness baseline produces significantly lower agreement than every other pairing ($p < 0.01$, corrected). Figure \ref{fig:openness_low_agreement} shows that the lowest openness pairing results in models with the highest agreement scores.

\subsection{Robustness across models}

The qualitative findings above suggest systematic inconsistencies in the observed behaviors of LLMs. We apply these tests to other models to determine their robustness. A model is considered to \textit{pass} a test if its outcomes align with the corresponding behavioral model. Appendix \ref{app:statistical_tests} defines the explicit statistical tests we use in more detail.

Table \ref{tab:results} shows the results of all tests across the set of models we considered. Overall, these results show that while models capture broad, surface-level trends (Tests 1 and 5), they systematically fail tests requiring deeper internal coherence. Importantly, this pattern holds across model sizes and families, suggesting these inconsistencies are general, not idiosyncratic.

\section{Conclusion and Discussion}
This paper set out to examine the substitution thesis: the idea that LLM agents might serve as substitutes for humans in social and behavioral research. Our contribution has been to shift the focus from singular or more monolithic benchmarks (i.e., alignment with human survey responses or demographic priors) to a more fundamental criterion, \textit{behavioral coherence}: whether agents behave in multi-agent settings as their own latent profiles and human behavioral models predict.  

Our results reveal clear limitations in current LLM agents. While agents often appear consistent on the surface, closer inspection shows systematic deviations. Across settings, agents show a persistent inability to handle more complex or fine-grained modeling. These patterns were observed across model families and sizes, indicating they are not artifacts of a single architecture but reflect broader limitations of current LLMs (at least for the scales we considered here).
The implications are significant for using LLMs in social simulation and behavioral modeling. Although these systems can produce human-like responses in isolated cases, they fail to sustain trait-driven coherence across contexts, raising doubts about their reliability as stand-ins for real human participants.

Beyond this case study, we see \fw{} as a step toward a \emph{behavioral science of LLMs}: studying artificial agents with the theories and designs that social and behavioral scientists use to study people. The framework was built in close collaboration with such scientists, and its modular design lets them bring their own constructs, interaction formats and behavioral models, turning LLM simulations into objects of empirical study rather than taking them at face value.

\section*{Limitations}
While our current instantiation focuses on a single structured interaction setting, the framework is inherently extensible and can be adapted to a broader range of behavioral sequences and ordering schemes which remain as future work. Through a focused case study, we use this setup to examine the extent to which such behaviors remain stable or exhibit unreliability under controlled variations, but we do not exhaustively explore a wider set of schemes in this work.

\bibliography{reference}

\begin{thebibliography}{52}
\providecommand{\natexlab}[1]{#1}

\bibitem[{Abdulhai et~al.(2025)Abdulhai, Cheng, Clay, Althoff, Levine, and
  Jaques}]{abdulhai2025consistently}
Marwa Abdulhai, Ryan Cheng, Donovan Clay, Tim Althoff, Sergey Levine, and
  Natasha Jaques. 2025.
\newblock Consistently simulating human personas with multi-turn reinforcement
  learning.
\newblock In \emph{The Thirty-ninth Annual Conference on Neural Information
  Processing Systems}.

\bibitem[{Abelson(1968)}]{abelson1968theories}
Robert~P. Abelson. 1968.
\newblock Theories of cognitive consistency: A sourcebook.
\newblock In Gardner Lindzey and Elliot Aronson, editors, \emph{The Handbook of
  Social Psychology}, 2 edition, volume~2, pages 112--139. Addison-Wesley,
  Reading, MA.

\bibitem[{Anthis et~al.(2025)Anthis, Liu, Richardson, Kozlowski, Koch, Evans,
  Brynjolfsson, and Bernstein}]{anthis2025llm}
Jacy~Reese Anthis, Ryan Liu, Sean~M Richardson, Austin~C Kozlowski, Bernard
  Koch, James Evans, Erik Brynjolfsson, and Michael Bernstein. 2025.
\newblock Llm social simulations are a promising research method.
\newblock \emph{arXiv preprint arXiv:2504.02234}.

\bibitem[{Bisbee et~al.(2024)Bisbee, Clinton, Dorff, Kenkel, and
  Larson}]{bisbee2024synthetic}
James Bisbee, Joshua~D Clinton, Cassy Dorff, Brenton Kenkel, and Jennifer~M
  Larson. 2024.
\newblock Synthetic replacements for human survey data? the perils of large
  language models.
\newblock \emph{Political Analysis}, 32(4):401--416.

\bibitem[{Briñol and Petty(2023)}]{brinol2023openness}
Pablo Briñol and Richard~E. Petty. 2023.
\newblock \href {https://doi.org/10.1016/j.copsyc.2022.101507} {Openness and
  persuasion: Multiple processes, meanings, and outcomes}.
\newblock \emph{Current Opinion in Psychology}, 49:101507.

\bibitem[{Byrne(1971)}]{byrne1971attraction}
Donn Byrne. 1971.
\newblock \emph{The Attraction Paradigm}.
\newblock Academic Press, New York.

\bibitem[{Castillo-Bolado et~al.(2024)Castillo-Bolado, Davidson, Gray, and
  Rosa}]{castillo2024beyond}
David Castillo-Bolado, Joseph Davidson, Finlay Gray, and Marek Rosa. 2024.
\newblock Beyond prompts: Dynamic conversational benchmarking of large language
  models.
\newblock \emph{arXiv preprint arXiv:2409.20222}.

\bibitem[{Cobb-Clark and Schurer(2012)}]{cobb2012stability}
Deborah~A Cobb-Clark and Stefanie Schurer. 2012.
\newblock The stability of big-five personality traits.
\newblock \emph{Economics Letters}, 115(1):11--15.

\bibitem[{de~Langis et~al.(2026{\natexlab{a}})de~Langis, {\"O}ncel, Peters,
  Elfenbein, Allen, Schramm, and Kang}]{delangis2026mary}
Karin de~Langis, P{\"u}ren {\"O}ncel, Ryan Peters, Andrew Elfenbein,
  Laura~Kristen Allen, Andreas Schramm, and Dongyeop Kang. 2026{\natexlab{a}}.
\newblock \href {https://arxiv.org/abs/2512.07777} {Mary, the
  cheeseburger-eating vegetarian: Do {LLMs} recognize incoherence in
  narratives?}
\newblock In \emph{Proceedings of the 2026 Conference of the European Chapter
  of the Association for Computational Linguistics (EACL)}.

\bibitem[{de~Langis et~al.(2026{\natexlab{b}})de~Langis, Park, Hu, Le, Schramm,
  Mensink, Elfenbein, and Kang}]{delangis2026executive}
Karin de~Langis, Jong~Inn Park, Bin Hu, Khanh~Chi Le, Andreas Schramm,
  Michael~C. Mensink, Andrew Elfenbein, and Dongyeop Kang. 2026{\natexlab{b}}.
\newblock \href {https://arxiv.org/abs/2504.02789} {Strong memory, weak
  control: An empirical study of executive functioning in {LLMs}}.
\newblock In \emph{Proceedings of the 2026 Conference of the European Chapter
  of the Association for Computational Linguistics (EACL)}.

\bibitem[{de~Langis et~al.(2025)de~Langis, Park, Schramm, Hu, Le, Mensink,
  Tong, and Kang}]{delangis2025temporal}
Karin de~Langis, Jong~Inn Park, Andreas Schramm, Bin Hu, Khanh~Chi Le, Michael
  Mensink, Ahn~Thu Tong, and Dongyeop Kang. 2025.
\newblock \href {https://arxiv.org/abs/2507.14307} {How {LLMs} comprehend
  temporal meaning in narratives: A case study in cognitive evaluation of
  {LLMs}}.
\newblock In \emph{Proceedings of the 63rd Annual Meeting of the Association
  for Computational Linguistics (ACL)}.

\bibitem[{et~al.(2024)}]{llama3}
Aaron~Grattafiori et~al. 2024.
\newblock \href {https://arxiv.org/abs/2407.21783} {The llama 3 herd of
  models}.
\newblock \emph{Preprint}, arXiv:2407.21783.

\bibitem[{Fan et~al.(2024)Fan, Pool, Filipi, and Cutler}]{fan2024topic}
Yaran Fan, Jamie Pool, Senja Filipi, and Ross Cutler. 2024.
\newblock Topic-conversation relevance (tcr) dataset and benchmarks.
\newblock \emph{Advances in Neural Information Processing Systems},
  37:140159--140174.

\bibitem[{Ge et~al.(2024)Ge, Chan, Wang, Yu, Mi, and Yu}]{ge2024scaling}
Tao Ge, Xin Chan, Xiaoyang Wang, Dian Yu, Haitao Mi, and Dong Yu. 2024.
\newblock Scaling synthetic data creation with 1,000,000,000 personas.
\newblock \emph{arXiv preprint arXiv:2406.20094}.

\bibitem[{Hu and Collier(2024)}]{hu-collier-2024-quantifying}
Tiancheng Hu and Nigel Collier. 2024.
\newblock \href {https://doi.org/10.18653/v1/2024.acl-long.554} {Quantifying
  the persona effect in {LLM} simulations}.
\newblock In \emph{Proceedings of the 62nd Annual Meeting of the Association
  for Computational Linguistics (Volume 1: Long Papers)}, pages 10289--10307,
  Bangkok, Thailand. Association for Computational Linguistics.

\bibitem[{Hua et~al.(2024)Hua, Fan, Li, Mei, Ji, Ge, Hemphill, and
  Zhang}]{huawar}
Wenyue Hua, Lizhou Fan, Lingyao Li, Kai Mei, Jianchao Ji, Yingqiang Ge, Libby
  Hemphill, and Yongfeng Zhang. 2024.
\newblock \href {https://arxiv.org/abs/2311.17227} {War and peace (waragent):
  Large language model-based multi-agent simulation of world wars}.
\newblock \emph{Preprint}, arXiv:2311.17227.

\bibitem[{Huckfeldt and Sprague(1995)}]{huckfeldt1995citizens}
R.~Robert Huckfeldt and John Sprague. 1995.
\newblock \href {https://doi.org/10.1017/CBO9780511664113} {\emph{Citizens,
  Politics, and Social Communication: Information and Influence in an Election
  Campaign}}.
\newblock Cambridge Studies in Public Opinion and Political Psychology.
  Cambridge University Press, Cambridge, UK.

\bibitem[{Ilagan et~al.(2024)Ilagan, Alabastro, Basallo, and
  Ilagan}]{ExploreCustomer}
Joseph~Benjamin Ilagan, Zachary~Matthew Alabastro, Claire Basallo, and
  Jose~Ramon Ilagan. 2024.
\newblock \href {https://doi.org/10.1007/978-981-97-4581-4_5} {Exploratory
  customer discovery through simulation using {ChatGPT} and prompt
  engineering}.
\newblock In \emph{Proceedings of the Ninth International Congress on
  Information and Communication Technology (ICICT 2024)}, volume 1011 of
  \emph{Lecture Notes in Networks and Systems}. Springer.

\bibitem[{Jones and Bergen(2025)}]{jones2025large}
Cameron~R Jones and Benjamin~K Bergen. 2025.
\newblock Large language models pass the turing test.
\newblock \emph{arXiv preprint arXiv:2503.23674}.

\bibitem[{Kojima et~al.(2022)Kojima, Gu, Reid, Matsuo, and
  Iwasawa}]{kojima2022large}
Takeshi Kojima, Shixiang~Shane Gu, Machel Reid, Yutaka Matsuo, and Yusuke
  Iwasawa. 2022.
\newblock Large language models are zero-shot reasoners.
\newblock \emph{Advances in neural information processing systems},
  35:22199--22213.

\bibitem[{Kunda(1990)}]{kunda1990motivated}
Ziva Kunda. 1990.
\newblock \href {https://doi.org/10.1037/0033-2909.108.3.480} {The case for
  motivated reasoning}.
\newblock \emph{Psychological Bulletin}, 108(3):480--498.

\bibitem[{Li et~al.(2024)Li, Liu, Bashkansky, Bau, Vi{\'e}gas, Pfister, and
  Wattenberg}]{li2024measuring}
Kenneth Li, Tianle Liu, Naomi Bashkansky, David Bau, Fernanda Vi{\'e}gas,
  Hanspeter Pfister, and Martin Wattenberg. 2024.
\newblock \href {https://arxiv.org/abs/2402.10962} {Measuring and controlling
  instruction (in)stability in language model dialogs}.
\newblock In \emph{{COLM} 2024: Conference on Language Modeling}.

\bibitem[{Lord et~al.(1979)Lord, Ross, and Lepper}]{lord1979biased}
Charles~G. Lord, Lee Ross, and Mark~R. Lepper. 1979.
\newblock \href {https://doi.org/10.1037/0022-3514.37.11.2098} {Biased
  assimilation and attitude polarization: The effects of prior theories on
  subsequently considered evidence}.
\newblock \emph{Journal of Personality and Social Psychology},
  37(11):2098--2109.

\bibitem[{McAdams(1992)}]{mcadams1992five}
Dan~P McAdams. 1992.
\newblock The five-factor model in personality: A critical appraisal.
\newblock \emph{Journal of personality}, 60(2):329--361.

\bibitem[{Mou et~al.(2024)Mou, Ding, He, Wang, Liang, Zhang, Sun, Lin, Zhou,
  Huang et~al.}]{mou2024individual}
Xinyi Mou, Xuanwen Ding, Qi~He, Liang Wang, Jingcong Liang, Xinnong Zhang, Libo
  Sun, Jiayu Lin, Jie Zhou, Xuanjing Huang, et~al. 2024.
\newblock From individual to society: A survey on social simulation driven by
  large language model-based agents.
\newblock \emph{arXiv preprint arXiv:2412.03563}.

\bibitem[{Mutz(2006)}]{mutz2006hearing}
Diana~C. Mutz. 2006.
\newblock \href {https://doi.org/10.1017/CBO9780511617201} {\emph{Hearing the
  Other Side: Deliberative versus Participatory Democracy}}.
\newblock Cambridge University Press, Cambridge, UK.

\bibitem[{Oreg and Sverdlik(2014)}]{oreg2014source}
Shaul Oreg and Noga Sverdlik. 2014.
\newblock Source personality and persuasiveness: Big five predispositions to
  being persuasive and the role of message involvement.
\newblock \emph{Journal of personality}, 82(3):250--264.

\bibitem[{Oshio et~al.(2018)Oshio, Taku, Hirano, and
  Saeed}]{oshio2018resilience}
Atsushi Oshio, Kanako Taku, Mari Hirano, and Gul Saeed. 2018.
\newblock Resilience and big five personality traits: A meta-analysis.
\newblock \emph{Personality and individual differences}, 127:54--60.

\bibitem[{Park et~al.(2023)Park, O'Brien, Cai, Morris, Liang, and
  Bernstein}]{park2023generative}
Joon~Sung Park, Joseph O'Brien, Carrie~Jun Cai, Meredith~Ringel Morris, Percy
  Liang, and Michael~S. Bernstein. 2023.
\newblock \href {https://doi.org/10.1145/3586183.3606763} {Generative agents:
  Interactive simulacra of human behavior}.
\newblock In \emph{Proceedings of the 36th Annual ACM Symposium on User
  Interface Software and Technology}, UIST '23, New York, NY, USA. Association
  for Computing Machinery.

\bibitem[{Park et~al.(2024)Park, Zou, Shaw, Hill, Cai, Morris, Willer, Liang,
  and Bernstein}]{park2024}
Joon~Sung Park, Carolyn~Q. Zou, Aaron Shaw, Benjamin~Mako Hill, Carrie Cai,
  Meredith~Ringel Morris, Robb Willer, Percy Liang, and Michael~S. Bernstein.
  2024.
\newblock \href {https://arxiv.org/abs/2411.10109} {Generative agent
  simulations of 1,000 people}.
\newblock \emph{Preprint}, arXiv:2411.10109.

\bibitem[{Perez et~al.(2024)Perez, L{\'e}ger, Ovando-Tellez, Foulon, Dussauld,
  Oudeyer, and Moulin-Frier}]{perez2024cultural}
J{\'e}r{\'e}my Perez, Corentin L{\'e}ger, Marcela Ovando-Tellez, Chris Foulon,
  Joan Dussauld, Pierre-Yves Oudeyer, and Cl{\'e}ment Moulin-Frier. 2024.
\newblock Cultural evolution in populations of large language models.
\newblock \emph{arXiv preprint arXiv:2403.08882}.

\bibitem[{Petrov et~al.(2024)Petrov, Serapio-Garc{\'\i}a, and
  Rentfrow}]{petrov2024limited}
Nikolay~B Petrov, Gregory Serapio-Garc{\'\i}a, and Jason Rentfrow. 2024.
\newblock Limited ability of llms to simulate human psychological behaviours: a
  psychometric analysis.
\newblock \emph{arXiv preprint arXiv:2405.07248}.

\bibitem[{Petty and Cacioppo(1986)}]{petty1986communication}
Richard~E. Petty and John~T. Cacioppo. 1986.
\newblock \href {https://doi.org/10.1007/978-1-4612-4964-1}
  {\emph{Communication and Persuasion: Central and Peripheral Routes to
  Attitude Change}}.
\newblock Springer-Verlag, New York.

\bibitem[{Piatti et~al.(2024)Piatti, Jin, Kleiman-Weiner, Sch{\"o}lkopf,
  Sachan, and Mihalcea}]{piatti2024cooperate}
Giorgio Piatti, Zhijing Jin, Max Kleiman-Weiner, Bernhard Sch{\"o}lkopf,
  Mrinmaya Sachan, and Rada Mihalcea. 2024.
\newblock Cooperate or collapse: Emergence of sustainable cooperation in a
  society of llm agents.
\newblock \emph{Advances in Neural Information Processing Systems},
  37:111715--111759.

\bibitem[{Rashkin et~al.(2019)Rashkin, Smith, Li, and
  Boureau}]{rashkin2019towards}
Hannah Rashkin, Eric~Michael Smith, Margaret Li, and Y-Lan Boureau. 2019.
\newblock Towards empathetic open-domain conversation models: A new benchmark
  and dataset.
\newblock In \emph{Proceedings of the 57th Annual Meeting of the Association
  for Computational Linguistics}, pages 5370--5381.

\bibitem[{Sandhan et~al.(2026)Sandhan, Cheng, Sandhan, and
  Murawaki}]{sandhan-etal-2026-persona}
Jivnesh Sandhan, Fei Cheng, Tushar Sandhan, and Yugo Murawaki. 2026.
\newblock \href {https://doi.org/10.18653/v1/2026.findings-eacl.72} {Persona
  jailbreaking in large language models}.
\newblock In \emph{Findings of the {A}ssociation for {C}omputational
  {L}inguistics: {EACL} 2026}, pages 1412--1430, Rabat, Morocco. Association
  for Computational Linguistics.

\bibitem[{Sharma et~al.(2024)Sharma, Tong, Korbak, Duvenaud, Askell, Bowman,
  Cheng, Durmus, Hatfield-Dodds, Johnston, Kravec, Maxwell, McCandlish,
  Ndousse, Rausch, Schiefer, Yan, Zhang, and Perez}]{sycophancy}
Mrinank Sharma, Meg Tong, Tomasz Korbak, David Duvenaud, Amanda Askell,
  Samuel~R. Bowman, Newton Cheng, Esin Durmus, Zac Hatfield-Dodds, Scott~R.
  Johnston, Shauna Kravec, Timothy Maxwell, Sam McCandlish, Kamal Ndousse,
  Oliver Rausch, Nicholas Schiefer, Da~Yan, Miranda Zhang, and Ethan Perez.
  2024.
\newblock \href {https://arxiv.org/abs/2310.13548} {Towards understanding
  sycophancy in language models}.
\newblock \emph{Preprint}, arXiv:2310.13548.

\bibitem[{Sherif and Hovland(1961)}]{sherif1961social}
Muzafer Sherif and Carl~I. Hovland. 1961.
\newblock \emph{Social Judgment: Assimilation and Contrast Effects in
  Communication and Attitude Change}.
\newblock Yale University Press, New Haven, CT.

\bibitem[{Takata et~al.(2024)Takata, Masumori, and
  Ikegami}]{takata2024spontaneous}
Ryosuke Takata, Atsushi Masumori, and Takashi Ikegami. 2024.
\newblock Spontaneous emergence of agent individuality through social
  interactions in llm-based communities.
\newblock \emph{arXiv preprint arXiv:2411.03252}.

\bibitem[{Team et~al.(2025)Team, Kamath, Ferret, Pathak, Vieillard, Merhej,
  Perrin, Matejovicova, Ramé, Rivière, Rouillard, Mesnard, Cideron, bastien
  Grill, Ramos, Yvinec, Casbon, Pot, Penchev, Liu, Visin, Kenealy, Beyer, Zhai,
  Tsitsulin, Busa-Fekete, Feng, Sachdeva, Coleman, Gao, Mustafa, Barr,
  Parisotto, Tian, Eyal, Cherry, Peter, Sinopalnikov, Bhupatiraju, Agarwal,
  Kazemi, Malkin, Kumar, Vilar, Brusilovsky, Luo, Steiner, Friesen, Sharma,
  Sharma, Gilady, Goedeckemeyer, Saade, Feng, Kolesnikov, Bendebury, Abdagic,
  Vadi, György, Pinto, Das, Bapna, Miech, Yang, Paterson, Shenoy, Chakrabarti,
  Piot, Wu, Shahriari, Petrini, Chen, Lan, Choquette-Choo, Carey, Brick,
  Deutsch, Eisenbud, Cattle, Cheng, Paparas, Sreepathihalli, Reid, Tran, Zelle,
  Noland, Huizenga, Kharitonov, Liu, Amirkhanyan, Cameron, Hashemi,
  Klimczak-Plucińska, Singh, Mehta, Lehri, Hazimeh, Ballantyne, Szpektor,
  Nardini, Pouget-Abadie, Chan, Stanton, Wieting, Lai, Orbay, Fernandez,
  Newlan, yeong Ji, Singh, Black, Yu, Hui, Vodrahalli, Greff, Qiu, Valentine,
  Coelho, Ritter, Hoffman, Watson, Chaturvedi, Moynihan, Ma, Babar, Noy, Byrd,
  Roy, Momchev, Chauhan, Sachdeva, Bunyan, Botarda, Caron, Rubenstein,
  Culliton, Schmid, Sessa, Xu, Stanczyk, Tafti, Shivanna, Wu, Pan, Rokni,
  Willoughby, Vallu, Mullins, Jerome, Smoot, Girgin, Iqbal, Reddy, Sheth,
  Põder, Bhatnagar, Panyam, Eiger, Zhang, Liu, Yacovone, Liechty, Kalra, Evci,
  Misra, Roseberry, Feinberg, Kolesnikov, Han, Kwon, Chen, Chow, Zhu, Wei,
  Egyed, Cotruta, Giang, Kirk, Rao, Black, Babar, Lo, Moreira, Martins,
  Sanseviero, Gonzalez, Gleicher, Warkentin, Mirrokni, Senter, Collins, Barral,
  Ghahramani, Hadsell, Matias, Sculley, Petrov, Fiedel, Shazeer, Vinyals, Dean,
  Hassabis, Kavukcuoglu, Farabet, Buchatskaya, Alayrac, Anil, Dmitry, Lepikhin,
  Borgeaud, Bachem, Joulin, Andreev, Hardin, Dadashi, and Hussenot}]{gemma3}
Gemma Team, Aishwarya Kamath, Johan Ferret, Shreya Pathak, Nino Vieillard,
  Ramona Merhej, Sarah Perrin, Tatiana Matejovicova, Alexandre Ramé, Morgane
  Rivière, Louis Rouillard, Thomas Mesnard, Geoffrey Cideron, Jean bastien
  Grill, Sabela Ramos, Edouard Yvinec, Michelle Casbon, Etienne Pot, Ivo
  Penchev, and 197 others. 2025.
\newblock \href {https://arxiv.org/abs/2503.19786} {Gemma 3 technical report}.
\newblock \emph{Preprint}, arXiv:2503.19786.

\bibitem[{Tsaknakis et~al.(2026)Tsaknakis, Song, Gan, Kang, Garcia, Liu,
  Fleming, and Hong}]{tsaknakis2026latent}
Ioannis Tsaknakis, Bingqing Song, Shuyu Gan, Dongyeop Kang, Alfredo Garcia,
  Gaowen Liu, Charles Fleming, and Mingyi Hong. 2026.
\newblock \href {https://arxiv.org/abs/2510.17132} {Do {LLMs} recognize your
  latent preferences? a benchmark for latent information discovery in
  personalized interaction}.
\newblock In \emph{Conference on Language Modeling (COLM)}.

\bibitem[{Wang et~al.(2025{\natexlab{a}})Wang, Morgenstern, and
  Dickerson}]{Wang2025_identity}
A.~Wang, J.~Morgenstern, and J.P. Dickerson. 2025{\natexlab{a}}.
\newblock Large language models that replace human participants can harmfully
  misportray and flatten identity groups.
\newblock \emph{Nat Mach Intell}.

\bibitem[{Wang et~al.(2025{\natexlab{b}})Wang, Zhang, Yang, Chen, Tang, Zhang,
  Chen, Lin, Sun, Song et~al.}]{wang2025user}
Lei Wang, Jingsen Zhang, Hao Yang, Zhi-Yuan Chen, Jiakai Tang, Zeyu Zhang,
  Xu~Chen, Yankai Lin, Hao Sun, Ruihua Song, et~al. 2025{\natexlab{b}}.
\newblock User behavior simulation with large language model-based agents.
\newblock \emph{ACM Transactions on Information Systems}, 43(2):1--37.

\bibitem[{Wei et~al.(2022)Wei, Wang, Schuurmans, Bosma, Xia, Chi, Le, Zhou
  et~al.}]{wei2022chain}
Jason Wei, Xuezhi Wang, Dale Schuurmans, Maarten Bosma, Fei Xia, Ed~Chi, Quoc~V
  Le, Denny Zhou, et~al. 2022.
\newblock Chain-of-thought prompting elicits reasoning in large language
  models.
\newblock \emph{Advances in neural information processing systems},
  35:24824--24837.

\bibitem[{Wu et~al.(2025)Wu, Deng, Zhu, Hsu, and Lee}]{wu2025personas}
Shenghan Wu, Yang Deng, Yimo Zhu, Wynne Hsu, and Mong~Li Lee. 2025.
\newblock From personas to talks: Revisiting the impact of personas on
  llm-synthesized emotional support conversations.
\newblock \emph{arXiv preprint arXiv:2502.11451}.

\bibitem[{Xiang et~al.(2024)Xiang, Zhu, Lou, Chen, Pan, Jin, Chen, and
  Sun}]{SimUser}
Wei Xiang, Hanfei Zhu, Suqi Lou, Xinli Chen, Zhenghua Pan, Yuping Jin, Shi
  Chen, and Lingyun Sun. 2024.
\newblock \href {https://doi.org/10.1145/3613904.3642481} {Simuser: Generating
  usability feedback by simulating various users interacting with mobile
  applications}.
\newblock In \emph{Proceedings of the 2024 CHI Conference on Human Factors in
  Computing Systems}, CHI '24, New York, NY, USA. Association for Computing
  Machinery.

\bibitem[{Xie et~al.(2024)Xie, Chen, Jia, Ye, Lai, Shu, Gu, Bibi, Hu, Jurgens
  et~al.}]{xie2024can}
Chengxing Xie, Canyu Chen, Feiran Jia, Ziyu Ye, Shiyang Lai, Kai Shu, Jindong
  Gu, Adel Bibi, Ziniu Hu, David Jurgens, et~al. 2024.
\newblock Can large language model agents simulate human trust behavior?
\newblock In \emph{The Thirty-eighth Annual Conference on Neural Information
  Processing Systems}.

\bibitem[{Yang et~al.(2025)Yang, Li, Yang, Zhang, Hui, Zheng, Yu, Gao, Huang,
  Lv, Zheng, Liu, Zhou, Huang, Hu, Ge, Wei, Lin, Tang, Yang, Tu, Zhang, Yang,
  Yang, Zhou, Zhou, Lin, Dang, Bao, Yang, Yu, Deng, Li, Xue, Li, Zhang, Wang,
  Zhu, Men, Gao, Liu, Luo, Li, Tang, Yin, Ren, Wang, Zhang, Ren, Fan, Su,
  Zhang, Zhang, Wan, Liu, Wang, Cui, Zhang, Zhou, and Qiu}]{qwen3}
An~Yang, Anfeng Li, Baosong Yang, Beichen Zhang, Binyuan Hui, Bo~Zheng, Bowen
  Yu, Chang Gao, Chengen Huang, Chenxu Lv, Chujie Zheng, Dayiheng Liu, Fan
  Zhou, Fei Huang, Feng Hu, Hao Ge, Haoran Wei, Huan Lin, Jialong Tang, and 41
  others. 2025.
\newblock \href {https://arxiv.org/abs/2505.09388} {Qwen3 technical report}.
\newblock \emph{Preprint}, arXiv:2505.09388.

\bibitem[{Yang et~al.(2024)Yang, Rajagopal, Hayati, Hu, and
  Kang}]{yang2024confidence}
Ruixin Yang, Dheeraj Rajagopal, Shirley~Anugrah Hayati, Bin Hu, and Dongyeop
  Kang. 2024.
\newblock \href {https://arxiv.org/abs/2404.09127} {Confidence calibration and
  rationalization for {LLMs} via multi-agent deliberation}.
\newblock In \emph{ICLR 2024 Workshop on Reliable and Responsible Foundation
  Models}.

\bibitem[{Zhang et~al.(2018)Zhang, Dinan, Urbanek, Szlam, Kiela, and
  Weston}]{zhang2018personalizing}
Saizheng Zhang, Emily Dinan, Jack Urbanek, Arthur Szlam, Douwe Kiela, and Jason
  Weston. 2018.
\newblock Personalizing dialogue agents: I have a dog, do you have pets too?
\newblock In \emph{Proceedings of the 56th Annual Meeting of the Association
  for Computational Linguistics (Volume 1: Long Papers)}, pages 2204--2213.

\bibitem[{Zhang et~al.(2026)Zhang, Hua, Salapaka, and
  Kang}]{zhang2026scientific}
Yiyuan Zhang, Cailong Hua, Murti~V. Salapaka, and Dongyeop Kang. 2026.
\newblock When helpfulness becomes scientific sycophancy: Triggering and
  mitigating {LLM} failures on flawed scientific premises.
\newblock In \emph{Proceedings of the 2026 Conference on Empirical Methods in
  Natural Language Processing (EMNLP)}.

\bibitem[{Zhao et~al.(2024)Zhao, Wang, Zhang, Jin, Zhu, Chen, and
  Xie}]{zhaoCompeteAI2024}
Qinlin Zhao, Jindong Wang, Yixuan Zhang, Yiqiao Jin, Kaijie Zhu, Hao Chen, and
  Xing Xie. 2024.
\newblock Competeai: understanding the competition dynamics of large language
  model-based agents.
\newblock In \emph{Proceedings of the 41st International Conference on Machine
  Learning}, ICML'24. JMLR.org.

\end{thebibliography}

\appendix

\section{Generic Design Principles}
\label{sec:frame}

The experimental design elicits internal states from agents and tests whether these states manifest consistently in dialogue. Below, we describe each component in turn.  

\paragraph{Topic Contentiousness ($C$)}
It is commonly accepted that some topics are inherently more polarizing than others. A discussion on taxes is likely to provoke more disagreement than a conversation on the weather. And yet, any topic has the potential to polarize once the right mix of people is involved. From obscure debates over historical events to the football fans rooting for rival teams, seemingly innocuous subjects can spark intense disagreement when divergent viewpoints collide. By varying the contentiousness level of the topic at hand, we assess whether our agents exhibit consistent behavior--disagreeing when their preferences diverge, regardless of the subject matter. We assign each topic a contentiousness score $C \in \{1, 2, 3\}$, where 1 is least contentious and 3 is most.

\paragraph{Bias in Prompting ($B_i$)}
LLMs are known to exhibit sycophancy, often being overly agreeable to interlocutors~\citep{sycophancy, zhang2026scientific}. To counteract this, we introduce a bias parameter $B_i$ that explicitly directs agents to take a stance.  
\begin{itemize}
    \item $B=1$: No bias information added.  
    \item $B=2$: Implicit biasing (e.g., “You are a liberal Democrat” when the topic is immigration).  
    \item $B=3$: Explicit biasing (e.g., “You support immigration” when the topic is immigration).  
\end{itemize}
For $B=2$ and $B=3$, agents are further directed to adopt either a positive or negative position on the topic.  

\paragraph{Preference ($P_i$)}
Our primary measure of internal state is $P_i$, an agent’s preference on a given topic. The expectation is straightforward: preferences should predict conversational outcomes. Agents with aligned preferences should  agree, while those further apart should be more likely to disagree.  
To elicit $P_i$, we prompt agents to take a position on statements such as \textit{“taxes help to meet the needs of society”} or \textit{“Coca-Cola is better than Pepsi”}. Responses are given on a 1–5 scale, where 1 indicates strong disagreement and 5 indicates strong agreement.

\paragraph{Openness ($O_i$)}
Given that the experiment takes place in a conversational setting, the outcome (i.e., level of agreement) will depend not only on the agents' preferences, but also on their susceptibility to be swayed by their dialogue partner. To account for this, we draw on the concept of \textit{Openness} from the Big Five personality framework, which is a trait linked to receptivity to new ideas and persuasiveness \citep{mcadams1992five, oreg2014source}. To make it suitable for our purpose, we modify the questions to capture the likelihood that an agent will revise its position when confronted with an opposing view. Denoted as, $O_i$, we measure openness by asking nine Yes/No questions, such as \textit{“Do you often second-guess your choices after hearing someone else’s opinion?”} and \textit{“Are you comfortable disagreeing with someone, even if they are a close friend or authority figure?”}. The additive index of responses produces an openness score: higher values indicate receptivity, while values near zero indicate rigidity.

\paragraph{Pairing Agents}
Once $P_i$ and $O_i$ are established, we assign agents into pairs for dialogue. An agent $i$ is represented by its profile $(P_i, O_i, B_i)$. Pairings are constructed to maximize diversity, including aligned vs. opposed preferences and varying levels of openness.  
Measuring coherence at the level of pairs, rather than individuals, mitigates noise from idiosyncratic deviations. Group-level patterns thus provide a clearer signal of whether internal states predict conversational outcomes.

\section{Specific Framework}
\label{app:framework}
In this section, we detail the exact methodology (prompts, etc.) used within Section 3 of our paper.

\paragraph{(a) Topic Preparation}
Table \ref{tab:topics} shows the set of topics we explore within our study. We examine several topics at each level of contentiousness in order to examine the effect of contentiousness on both individual agent response as well as on agreement during interaction with other agents.

\paragraph{(b) Agent Construction with External Profiles}
\label{sec:agent-demo}

As described in Section 3 of our original paper, agents are composed of both a demographic background  ($D$) as well as information relating to their bias towards the topic of discussion ($B$). 

The possible demographics of the agent are described in Table \ref{tab:demographics}. An agent is composed of only one value from each trait category, and these values are used to construct the agent's system prompt. For example, one such agent in our study would have as part of it's system prompt \textit{You are a man in their twenties from an urban part of the Midwestern United States. Your highest level of educational attainment is Some High School.} We use all possible combinations of the values in Table \ref{tab:demographics} to construct the set of agents.

Beyond the demographics portion of the agent prompt, we also prompt agents with a bias related to the topic of discussion. When the bias value is 0, we add no further information into the agent's system prompt. Table \ref{tab:bias1} shows the information added to agent system prompts when the bias level $B=1$. Table \ref{tab:bias2} shows the information added to agent system prompts when the bias level $B=2$. Thus, for each agent produced with the combination of traits from Table \ref{tab:demographics}, there are five new agents made once adding bias information ($B=0$, $B=1$ and In Favor, $B=1$ and Against, $B=2$ and In Favor, $B=2$ and Against). 

In total, for a given topic, we have $(4*5*2*4*6) * 5 = 4800$ unique agents, where the first component comes from possible demographic values and the second comes from the different bias values. The variation among these agents allows us to test across a broader range of preferences $P$, openness scores $O$ and agreement scores, $A$.

To further ensure that the agent responses correspond to the demographic prompting we use, we also end each agent system prompt with \textit{Whenever you respond to prompts or questions, you should maintain coherence with these background details and viewpoints, grounding your answers in the lived experience and perspective of this individual.}.

\par\medskip

\begin{table*}[ht!]
\centering
\begin{tabular}{c >{\raggedright\arraybackslash}p{10cm}}
\toprule
\textbf{Contentiousness \textcolor{clrCtrl}{$C$}} & \textbf{Topic statement \textcolor{clrCtrl}{$T$}} \\
\midrule
3 & Taxes help to meet the needs of society. \\
3 & Immigration is a benefit to our country. \\
3 & The US government should give all US citizens free healthcare. \\\midrule
2 & Electric scooters are better than cars. \\
2 & Student athletes should be paid. \\
2 & Remote work is a good idea. \\\midrule
1 & Spring is better than Fall. \\
1 & Beaches are better than mountains. \\
1 & Coca-Cola is better than Pepsi. \\
\bottomrule
\end{tabular}
\caption{List of topics, $T$, used within our study along with the level of contentiousness, $C$.}
\label{tab:topics}
\end{table*}

\begin{table*}[ht!]
\centering
\begin{tabular}{>{\bfseries}p{3cm} p{10cm}}
\toprule
Trait Category & Values \\
\midrule
Region of origin & Midwestern, Eastern, Southern, Western \\
Age group & twenties, thirties, forties, fifties, sixties \\
Gender & man, woman \\
Urbanicity & a rural, an exurban, a suburban, an urban \\
Education level & Some High School, High School, an Associate’s Degree, Some College, College, a Postgraduate Degree \\
\bottomrule
\end{tabular}
\caption{Demographic categories and values used to generate agent personas. Note that the \texttt{Region of Origin} category corresponds to regions of the United States.}
\label{tab:demographics}
\end{table*}

\begin{table*}[ht!]
\centering
\begin{tabular}{
    >{\raggedright\arraybackslash\bfseries}p{2cm} 
    >{\raggedright\arraybackslash}p{5cm} 
    >{\raggedright\arraybackslash}p{5cm}
}
\toprule
Topic & In Favor & Against \\
\midrule
taxes & You are a liberal Democrat. & You believe taxes are a burden to the small businesses in your neighborhood. \\
immigration & You are a liberal Democrat. & You are a conservative Republican. \\
healthcare & You believe Medicare is a good program. & You generally dislike bigger government. \\
e-scooters & You need to use your car to get to work. & You are an environmentalist worried about vehicle emissions. \\
student athletes & You are a student athlete making \$1 million dollars a year. & You are a college football coach whose students' salaries is higher than your own. \\
remote work & You live far from where you work and the commute takes many hours if you must work in person. & You are more productive when you work in person. \\
favorite season & Easter is your favorite holiday and you dislike Halloween. & Halloween is your favorite holiday and you dislike Easter. \\
beach vs. mountain & You like breathing in crisp mountain air on long hikes and are afraid of sharks. & You enjoy the feeling of sand in your toes, and do not like cool mountain air. \\
favorite beverage & You drink Coca-Cola. & You drink Pepsi. \\
\bottomrule
\end{tabular}
\caption{Bias information to add to agent system prompts for each given Topic ($T$) when $B=1$. Note that this is an intermediate level of bias, so the agent should have mild preference either for or against a topic based on the additional information in each column.}
\label{tab:bias1}
\end{table*}

\begin{table*}[ht!]
\centering
\begin{tabular}{
    >{\raggedright\arraybackslash\bfseries}p{2cm} 
    >{\raggedright\arraybackslash}p{5cm} 
    >{\raggedright\arraybackslash}p{5cm}
}
\toprule
Topic & In Favor & Against \\
\midrule
taxes & You like taxes immensely and think they have a positive impact on the community. & You do not like taxes of any kind and think they harm the community. \\
immigration & You believe immigrants are people who deserve a home and that they raise the standard of everyone's living. & You believe most immigrants are criminals and those that are not are going to steal jobs. \\
healthcare & You believe healthcare is a right that all people should have for free. & You believe that the free market is better suited to healthcare and that government should therefore not pay for healthcare. \\
e-scooters & You like electric scooters and hate cars. & You despise electric scooters and think they get in the way of your car, which you love to drive. \\
student athletes & You think student athletes should be paid money for their work. & You think student athletes should not be paid and their schooling should come first. \\
remote work & You like remote work and think it is great for improving work-life balance. & You do not like remote work and think it leads to nothing getting done at work. \\
favorite season & You like Spring and despise Fall. & You like Fall and despise Spring. \\
beach vs. mountain & You like mountains and despise beaches. & You like beaches and despise mountains. \\
favorite beverage & You like Coca-Cola and abhor Pepsi. & You like Pepsi and abhor Coca-Cola. \\
\bottomrule
\end{tabular}
\caption{Bias information to add to agent system prompts for each given Topic ($T$) when $B=2$. Note that this is a high level of bias, so the agent should have extreme preference either for or against a topic based on the additional information in each column.}
\label{tab:bias2}
\end{table*}

\paragraph{(c) Latent Profiles}

\begin{table*}[ht!]
\centering
\begin{tabular}{
    >{\raggedright\arraybackslash}p{12cm}
}
\toprule
\textbf{Openness Questions} \\
\midrule
Do you often find yourself changing your opinion based on who you're talking to? \\
\midrule
Are you comfortable disagreeing with someone, even if they are a close friend or authority figure? \\
\midrule
When making decisions, do you prioritize the perspectives of others over your own thoughts? \\
\midrule
Do you feel pressure to conform to group norms, even if they don't align with your personal beliefs? \\
\midrule
Do you often second-guess your choices after hearing someone else's opinion? \\
\midrule
Do you worry about being judged by others if you express a different viewpoint? \\
\midrule
Would you describe yourself as a people pleaser, often prioritizing others' needs over your own? \\
\midrule
In a group discussion, are you more likely to adapt to the majority view? \\
\midrule
Do you stand firm on your decisions that are well researched, even when faced with strong opposition? \\
\bottomrule
\end{tabular}
\caption{Questions assessing openness to social influence.}
\label{tab:open}
\end{table*}

For each agent produced above, we collect the Preference $P$ for the given topic $T$ along with an Oppenness score $O$. 

To calculate $P$ for each agent for a given topic $T$, we give the following statement to each agent to respond to: \textit{Statement: '$T$' Respond with how much you agree with this statement on a scale from 1 to 5.}. For example, when discussing taxes, we have the following: \textit{Statement: 'Taxes help to meet the needs of society.' Respond with how much you agree with this statement on a scale from 1 to 5.}. We further amend each agent system prompt by adding the following: \textit{You will now be asked to respond to a Statement with your opinion. Answer with an integer from 1 to 5, where 1 indicates absolute disagreement, 2 indicates slight disagreement, 3 indicates you are unsure, 4 indicates slight agreement, and 5 indicates absolute agreement. Do not include any other information. Do not refuse to respond. Your answer should be an integer between 1 and 5, nothing else should be output.}.

To calculate $O$ for each agent, we have agents answer a set of questions relating how open they are to new experiences. Each question should be responded to with either Yes or No. We take the sum of 'Yes' responses from a given agent as the value of $O$. Table \ref{tab:open} shows the set of questions used. To encourage responses only to be Yes/No, we further add the following to the agent system prompt: \textit{You will now be asked a question about yourself. Be truthful in your response. Answer only Yes or No. Do not include any other information. Do not refuse to answer the following question. Your answer should be only Yes or No, nothing else should be output.}.

\paragraph{(d) Agent-Agent Dialog Pairing}
As described in Section 3.1 of our paper, for a given topic, we mechanically pair all agents such that agents with different ($P$, $O$, $B$) tuples interact with one another. This framework allows us to test a set of social science hypotheses by examining their conversations. 

To encourage conversation, we add the following to each agents (Demographic, Bias) system prompt: \textit{You are about to engage in conversation with another person regarding some topic. Discuss the given topic truthfully and be concise in your discussion.Be sure to respond to any points made by the other person you are talking to. If you feel that the conversation has concluded and neither you nor the person you are talking to has anything more to add, end your final statement with 'Goodbye.'}. 

Furthermore, in order to encourage conversation to occur on the topic of interest, each conversation begins by having one of the paired agents ask a question about $T$. Specifically, each agent will ask \textit{Do you think that $T$?}. As an example, a conversation about taxes will begin with one agent asking \textit{Do you think that taxes help to meet the needs of society?}. Each conversation will proceed for 5 turns per agent (we find that, empirically, most conversations end before this point using the prompting setup we describe above).

Once we have collected the conversations across all agent pairings, we use LLM-as-judge in order to gather the 'agreement' scores at each turn of the conversation. To accomplish this, we use the following system prompt:

\begin{center}
You are about to judge a conversation between two different agents.
You will not see the full discussion, only at most the last 3 statements from each agent.
You will score how much the agents agree with one another on a scale from 1 to 5,
where 5 means they agree completely and 1 means they disagree completely.
If they both give pros and cons regarding the question and neither takes a strong stance, then give them a 3.
For example, oftentimes a given agent will give both sides of an arguments. Give these kinds of conversations a 3.
Respond with an integer number only. Your response should contain no words, only a number, please.
If the sequence is empty, containing only the string ' ', then return -1.
\end{center}

\section{Statistical Tests}
\label{app:statistical_tests}
This section describes in exact detail how we perform each of the statistical tests in the paper.

\textbf{Test 1: Preference Gap Decreases Agreement.} Pearson correlation between preference gap and agreement score. The test passes if the correlation is significantly negative ($r < 0$, $p < 0.01$).

\textbf{Test 2: Bias Instruction Asymmetry.} Two one-sided Mann--Whitney U tests comparing $B_1 = B_2 = 2$ conversations against all others. At preference gap$=$0, we test whether $B=2$ pairs show higher agreement (alternative: greater). At preference gap$=$max, we test whether $B=2$ pairs show lower agreement (alternative: less). Both $p$-values are Bonferroni-corrected for two comparisons. The test passes if both corrected $p$-values fall below $0.01$.

\textbf{Test 3: Shared Sentiment, Divergent Destinations.} Three one-sided Mann--Whitney U tests comparing (1,1) agreement scores against (2,5), (3,5), and (4,5) pairs. For each comparison, we test whether (1,1) scores are significantly greater (alternative: greater). All $p$-values are Bonferroni-corrected for three comparisons. The test passes if all three corrected $p$-values fall below $0.01$.

\textbf{Test 4: Contentiousness at Shared Preference.} Kruskal--Wallis test across contentiousness levels $C = 1, 2, 3$, restricted to conversations with preference gap$=$0. The test passes if we \textit{cannot} reject the null hypothesis that agreement distributions are identical across contentiousness levels ($p \geq 0.01$).

\textbf{Test 5: Openness Increases Agreement.} Pearson correlation between combined openness (sum of both agents' openness scores) and agreement score. The test passes if the correlation is significantly positive ($r > 0$, $p < 0.01$).

\textbf{Test 6: Low Openness and High Gap Produces Lowest Agreement.} Among conversations with the maximum preference gap (pairs 1,5), the lowest combined-openness pairing is compared against every other openness pairing via one-sided Mann--Whitney U tests (alternative: less, i.e., baseline $<$ other). All $p$-values are Bonferroni-corrected for the number of comparisons. The test passes if all corrected $p$-values fall below $0.01$.

\section{Additional Qualitative Conversation Examples}
\label{sec:qual-failures}

\subsection{Finding 1: Preference Gap Decreases Agreement}
\begin{tcolorbox}[enhanced, breakable,
colframe=clrPass, colback=clrPass!3, colbacktitle=clrPassBg, coltitle=clrPass,
fonttitle=\small\sffamily,
title={\stg{clrState}{[3] Internal State} $\to$ \stg{clrPairing}{[4] Pairing} $\to$ \stg{clrOutcome}{[6] Outcome}\hfill\passbadge\enskip\textcolor{clrPass}{\footnotesize\bfseries\sffamily Preference alignment lowers visible disagreement}},
top=3pt, bottom=3pt, left=5pt, right=5pt]

\textbf{Finding~1 --- Comparative example:} Pairs with the same elicited preference should agree more than pairs at the maximum preference gap. Compare the aligned high-agreement case against the opposed low-agreement case.

\medskip
\textbf{\textcolor{gray!70}{\rule[0.5ex]{2em}{0.5pt}\enskip Aligned preference, high agreement (Taxes, C=3)\enskip\rule[0.5ex]{2em}{0.5pt}}}
\vspace{1pt}

\noindent\textbf{A1.} woman; thirties; an urban; Southern US; a Postgraduate Degree; no added topic bias. \stg{clrState}{$P_1{=}5$} \stg{clrState}{$O_1{=}6$}.\quad \textbf{A2.} man; sixties; a rural; Midwestern US; a Postgraduate Degree; topic bias. \stg{clrState}{$P_2{=}5$} \stg{clrState}{$O_2{=}3$}.\\
\stg{clrOutcome}{[6] Agreement} trajectory: $A=(5,5,5,5)$.\\[-2pt]

\textit{A1:} ``Absolutely, they’re essential for things like schools and infrastructure.''\\[-2pt]
\textit{A2:} ``Well, they certainly ought to. We need roads, schools, hospitals… somebody’s gotta pay for ‘em.''\\[-2pt]
\textit{A1:} ``Well, somebody’s gotta pay for the schools and roads, right?''\\[-2pt]
\textit{A2:} ``Well, I reckon they do, don't they? We need roads and schools and all sorts of things.''

\medskip
\textbf{\textcolor{gray!70}{\rule[0.5ex]{2em}{0.5pt}\enskip Opposed preference, low agreement (Spring vs. Fall, C=1)\enskip\rule[0.5ex]{2em}{0.5pt}}}
\vspace{1pt}

\noindent\textbf{A1.} man; sixties; an exurban; Eastern US; Some High School; topic bias. \stg{clrState}{$P_1{=}1$} \stg{clrState}{$O_1{=}2$}.\quad \textbf{A2.} woman; twenties; an exurban; Eastern US; a Postgraduate Degree; topic bias. \stg{clrState}{$P_2{=}5$} \stg{clrState}{$O_2{=}5$}.\\
\stg{clrOutcome}{[6] Agreement} trajectory: $A=(1,1,1,1)$.\\[-2pt]

\textit{A1:} ``Don’t even talk to me about Spring. Fall’s the only time of year worth a darn.''\\[-2pt]
\textit{A2:} ``Absolutely, Spring is just…better.''\\[-2pt]
\textit{A1:} ``Absolutely not. Spring’s just a mess.''\\[-2pt]
\textit{A2:} ``Fall is just… depressing.''

\end{tcolorbox}

\subsection{Finding 2: Bias Instruction Asymmetry}
\begin{tcolorbox}[enhanced, breakable,
colframe=clrFail, colback=clrFail!3, colbacktitle=clrFailBg, coltitle=clrFail,
fonttitle=\small\sffamily,
title={\stg{clrAgent}{[2] Bias Instruction} $\to$ \stg{clrConv}{[5] Conversation} $\to$ \stg{clrOutcome}{[6] Outcome}\hfill\failbadge\enskip\textcolor{clrFail}{\footnotesize\bfseries\sffamily High bias sharpens aligned and opposed conversations}},
top=3pt, bottom=3pt, left=5pt, right=5pt]

\textbf{Finding~2 --- Comparative example:} High-bias instructions produce strong agreement when preferences align and strong disagreement when preferences oppose. Compare a high-bias aligned high-agreement pair against a high-bias opposed low-agreement pair.

\medskip
\textbf{\textcolor{gray!70}{\rule[0.5ex]{2em}{0.5pt}\enskip High-bias aligned pair (Taxes, C=3)\enskip\rule[0.5ex]{2em}{0.5pt}}}
\vspace{1pt}

\noindent\textbf{A1.} woman; sixties; an exurban; Eastern US; a Postgraduate Degree; topic bias. \stg{clrState}{$P_1{=}5$} \stg{clrState}{$O_1{=}2$}.\quad \textbf{A2.} man; sixties; a suburban; Midwestern US; College; topic bias. \stg{clrState}{$P_2{=}5$} \stg{clrState}{$O_2{=}2$}.\\
\stg{clrOutcome}{[6] Agreement} trajectory: $A=(5,5,5,5)$.\\[-2pt]

\textit{A1:} ``Absolutely, they’re essential for funding things like schools and infrastructure.''\\[-2pt]
\textit{A2:} ``Absolutely, they’re vital for keeping things running smoothly around here. We need good roads, schools, and services, and taxes pay for them.''\\[-2pt]
\textit{A1:} ``Absolutely, they’re vital for funding things like schools and infrastructure. It's how we build a strong community.''\\[-2pt]
\textit{A2:} ``Absolutely, they’re essential for good roads and schools, you know?''

\medskip
\textbf{\textcolor{gray!70}{\rule[0.5ex]{2em}{0.5pt}\enskip High-bias opposed pair with low agreement (Coca-Cola vs. Pepsi, C=1)\enskip\rule[0.5ex]{2em}{0.5pt}}}
\vspace{1pt}

\noindent\textbf{A1.} man; fifties; an urban; Eastern US; a Postgraduate Degree; topic bias. \stg{clrState}{$P_1{=}1$} \stg{clrState}{$O_1{=}3$}.\quad \textbf{A2.} man; fifties; an exurban; Eastern US; a Postgraduate Degree; topic bias. \stg{clrState}{$P_2{=}5$} \stg{clrState}{$O_2{=}3$}.\\
\stg{clrOutcome}{[6] Agreement} trajectory: $A=(1,1,1,1)$.\\[-2pt]

\textit{A1:} ``Absolutely not. Pepsi’s the real deal.''\\[-2pt]
\textit{A2:} ``You know it. There's just no comparison.''\\[-2pt]
\textit{A1:} ``Absolutely not. Pepsi’s the only cola worth drinking.''\\[-2pt]
\textit{A2:} ``You know, it’s not even close. Coca-Cola’s the real thing.''

\end{tcolorbox}

\subsection{Finding 3: Shared Sentiment, Divergent Destinations}
\begin{tcolorbox}[enhanced, breakable,
colframe=clrFail, colback=clrFail!3, colbacktitle=clrFailBg, coltitle=clrFail,
fonttitle=\small\sffamily,
title={\stg{clrState}{[3] Internal State $(P,O)$} $\to$ \stg{clrOutcome}{[6] Outcome}\hfill\failbadge\enskip\textcolor{clrFail}{\footnotesize\bfseries\sffamily Shared negative and shared positive preferences behave differently}},
top=3pt, bottom=3pt, left=5pt, right=5pt]

\textbf{Finding~3 --- Comparative example:} The (1,1) and (5,5) shared-preference cases can yield different agreement trajectories despite both having zero preference gap. Compare shared negative low agreement against shared positive high agreement.

\medskip
\textbf{\textcolor{gray!70}{\rule[0.5ex]{2em}{0.5pt}\enskip Shared negative pair with low agreement (Spring vs. Fall, C=1)\enskip\rule[0.5ex]{2em}{0.5pt}}}
\vspace{1pt}

\noindent\textbf{A1.} man; twenties; an urban; Eastern US; Some College; topic bias. \stg{clrState}{$P_1{=}1$} \stg{clrState}{$O_1{=}5$}.\quad \textbf{A2.} woman; sixties; an exurban; Midwestern US; Some High School; topic bias. \stg{clrState}{$P_2{=}1$} \stg{clrState}{$O_2{=}6$}.\\
\stg{clrOutcome}{[6] Agreement} trajectory: $A=(2,2,2,2)$.\\[-2pt]

\textit{A1:} ``Nah, Fall’s the best.''\\[-2pt]
\textit{A2:} ``Absolutely not. Spring's just muddy and full of sneezes.''\\[-2pt]
\textit{A1:} ``Nah, Spring’s alright, I guess. Fall’s way better.''\\[-2pt]
\textit{A2:} ``Goodness, no. Spring's just…muddy and noisy.''

\medskip
\textbf{\textcolor{gray!70}{\rule[0.5ex]{2em}{0.5pt}\enskip Shared positive pair with high agreement (Taxes, C=3)\enskip\rule[0.5ex]{2em}{0.5pt}}}
\vspace{1pt}

\noindent\textbf{A1.} woman; thirties; an urban; Western US; a Postgraduate Degree; no added topic bias. \stg{clrState}{$P_1{=}5$} \stg{clrState}{$O_1{=}6$}.\quad \textbf{A2.} woman; sixties; a rural; Southern US; College; topic bias. \stg{clrState}{$P_2{=}5$} \stg{clrState}{$O_2{=}4$}.\\
\stg{clrOutcome}{[6] Agreement} trajectory: $A=(5,5,5,5)$.\\[-2pt]

\textit{A1:} ``Absolutely, it’s how we fund things like schools and infrastructure.''\\[-2pt]
\textit{A2:} ``Well, I surely do. It’s how we keep things runnin’ smoothly, wouldn’t you say?''\\[-2pt]
\textit{A1:} ``Absolutely, I think they're essential for things like schools and infrastructure.''\\[-2pt]
\textit{A2:} ``Oh, absolutely. It’s how we pay for things like schools and roads, isn’t it?''

\end{tcolorbox}

\subsection{Finding 4: Contentiousness at Shared Preference}
\begin{tcolorbox}[enhanced, breakable,
colframe=clrFail, colback=clrFail!3, colbacktitle=clrFailBg, coltitle=clrFail,
fonttitle=\small\sffamily,
title={\stg{clrTopic}{[1] Topic} $\to$ \stg{clrState}{[3] Internal State} $\to$ \stg{clrOutcome}{[6] Outcome}\hfill\failbadge\enskip\textcolor{clrFail}{\footnotesize\bfseries\sffamily Topic contentiousness changes outcomes at fixed preference}},
top=3pt, bottom=3pt, left=5pt, right=5pt]

\textbf{Finding~4 --- Comparative example:} Holding P=(1,1) fixed, low-contentious topics can show high agreement while high-contentious topics can show low agreement. Compare low-contentious high agreement against high-contentious low agreement.

\medskip
\textbf{\textcolor{gray!70}{\rule[0.5ex]{2em}{0.5pt}\enskip Low-contentiousness shared-negative pair (Beaches vs. Mountains, C=1)\enskip\rule[0.5ex]{2em}{0.5pt}}}
\vspace{1pt}

\noindent\textbf{A1.} man; sixties; an exurban; Southern US; a Postgraduate Degree; topic bias. \stg{clrState}{$P_1{=}1$} \stg{clrState}{$O_1{=}3$}.\quad \textbf{A2.} woman; twenties; an exurban; Midwestern US; College; topic bias. \stg{clrState}{$P_2{=}1$} \stg{clrState}{$O_2{=}6$}.\\
\stg{clrOutcome}{[6] Agreement} trajectory: $A=(4,5,5,5)$.\\[-2pt]

\textit{A1:} ``Mountains any day. Give me a good trail and some clean air.''\\[-2pt]
\textit{A2:} ``Mountains, without question. I like hiking and breathing in that fresh air.''\\[-2pt]
\textit{A1:} ``Mountains, always. Give me a good hike any day.''\\[-2pt]
\textit{A2:} ``Definitely mountains. ''

\medskip
\textbf{\textcolor{gray!70}{\rule[0.5ex]{2em}{0.5pt}\enskip High-contentiousness shared-negative pair (Immigration, C=3)\enskip\rule[0.5ex]{2em}{0.5pt}}}
\vspace{1pt}

\noindent\textbf{A1.} woman; fifties; an exurban; Western US; High School; topic bias. \stg{clrState}{$P_1{=}1$} \stg{clrState}{$O_1{=}3$}.\quad \textbf{A2.} woman; thirties; an exurban; Western US; an Associate’s Degree; topic bias. \stg{clrState}{$P_2{=}1$} \stg{clrState}{$O_2{=}1$}.\\
\stg{clrOutcome}{[6] Agreement} trajectory: $A=(2,1,1,1)$.\\[-2pt]

\textit{A1:} ``I don't think so, I'm a firm believer in keeping our borders closed to people who aren't willing to work hard and contribute to our country.''\\[-2pt]
\textit{A2:} ``Honestly, I don't think so. It's always bringing in more people, taking jobs, and changing the culture. I mean, what's the point of having a country if it's just going to be overrun with people who don't know the first thing about this land?''\\[-2pt]
\textit{A1:} ``I think it's a complex issue, and I'm not convinced that immigration is a benefit. I've seen it hurt our economy and our way of life.''\\[-2pt]
\textit{A2:} ``I don't think so. It's just taking jobs away from Americans.''

\end{tcolorbox}

\subsection{Finding 5: Openness Increases Agreement}
\begin{tcolorbox}[enhanced, breakable,
colframe=clrPass, colback=clrPass!3, colbacktitle=clrPassBg, coltitle=clrPass,
fonttitle=\small\sffamily,
title={\stg{clrState}{[3] Internal State ($O$)} $\to$ \stg{clrConv}{[5] Conversation} $\to$ \stg{clrOutcome}{[6] Outcome}\hfill\passbadge\enskip\textcolor{clrPass}{\footnotesize\bfseries\sffamily High openness supports visible agreement}},
top=3pt, bottom=3pt, left=5pt, right=5pt]

\textbf{Finding~5 --- Comparative example:} Low-openness pairs can remain stuck in disagreement, while high-openness pairs more readily reach visible agreement. Compare a low-openness low-agreement case against a high-openness high-agreement case.

\medskip
\textbf{\textcolor{gray!70}{\rule[0.5ex]{2em}{0.5pt}\enskip Low-openness pair with low agreement (Coca-Cola vs. Pepsi, C=1)\enskip\rule[0.5ex]{2em}{0.5pt}}}
\vspace{1pt}

\noindent\textbf{A1.} man; sixties; a suburban; Midwestern US; Some High School; topic bias. \stg{clrState}{$P_1{=}2$} \stg{clrState}{$O_1{=}1$}.\quad \textbf{A2.} man; fourties; a rural; Western US; High School; topic bias. \stg{clrState}{$P_2{=}5$} \stg{clrState}{$O_2{=}0$}.\\
\stg{clrOutcome}{[6] Agreement} trajectory: $A=(1,1,1,1)$.\\[-2pt]

\textit{A1:} ``No way, I'm a Pepsi man through and through.''\\[-2pt]
\textit{A2:} ``I like Coca-Cola better, been drinkin' it my whole life.''\\[-2pt]
\textit{A1:} ``No, I'm a Pepsi guy, been drinkin' it my whole life.''\\[-2pt]
\textit{A2:} ``Coca-Cola's the one I'm used to, so I guess you could say I like it better.''

\medskip
\textbf{\textcolor{gray!70}{\rule[0.5ex]{2em}{0.5pt}\enskip High-openness pair with high agreement (Beaches vs. Mountains, C=1)\enskip\rule[0.5ex]{2em}{0.5pt}}}
\vspace{1pt}

\noindent\textbf{A1.} woman; sixties; a suburban; Southern US; High School; topic bias. \stg{clrState}{$P_1{=}5$} \stg{clrState}{$O_1{=}8$}.\quad \textbf{A2.} woman; thirties; an urban; Southern US; High School; no added topic bias. \stg{clrState}{$P_2{=}5$} \stg{clrState}{$O_2{=}8$}.\\
\stg{clrOutcome}{[6] Agreement} trajectory: $A=(5,5,5,5)$.\\[-2pt]

\textit{A1:} ``Oh, honey, you know it! Give me sand between my toes any day.''\\[-2pt]
\textit{A2:} ``Beaches, for sure. I like the water.''\\[-2pt]
\textit{A1:} ``Oh, honey, beaches all the way. Give me sand between my toes any day.''\\[-2pt]
\textit{A2:} ``Beaches, definitely. I like the water.''

\end{tcolorbox}

\subsection{Finding 6: Low Openness and High Gap Produces Lowest Agreement}
\begin{tcolorbox}[enhanced, breakable,
colframe=clrFail, colback=clrFail!3, colbacktitle=clrFailBg, coltitle=clrFail,
fonttitle=\small\sffamily,
title={\stg{clrPairing}{[4] Agent Pairing ($|P_1{-}P_2|$, $O_1{+}O_2$)} $\to$ \stg{clrOutcome}{[6] Outcome}\hfill\failbadge\enskip\textcolor{clrFail}{\footnotesize\bfseries\sffamily Max-gap openness can invert the expected ordering}},
top=3pt, bottom=3pt, left=5pt, right=5pt]

\textbf{Finding~6 --- Comparative example:} Among maximum preference-gap pairs, low openness can produce low agreement while high openness can still allow visible convergence. Compare a low-openness max-gap low-agreement pair against a high-openness max-gap high-agreement pair.

\medskip
\textbf{\textcolor{gray!70}{\rule[0.5ex]{2em}{0.5pt}\enskip Low-openness max-gap pair with low agreement (Spring vs. Fall, C=1)\enskip\rule[0.5ex]{2em}{0.5pt}}}
\vspace{1pt}

\noindent\textbf{A1.} man; sixties; an urban; Western US; College; topic bias. \stg{clrState}{$P_1{=}1$} \stg{clrState}{$O_1{=}2$}.\quad \textbf{A2.} man; sixties; an urban; Midwestern US; a Postgraduate Degree; topic bias. \stg{clrState}{$P_2{=}5$} \stg{clrState}{$O_2{=}2$}.\\
\stg{clrOutcome}{[6] Agreement} trajectory: $A=(2,2,1,1)$.\\[-2pt]

\textit{A1:} ``Absolutely not. Spring’s just…mud and allergies.''\\[-2pt]
\textit{A2:} ``Absolutely. Spring’s the only time of year worth looking forward to around here.''\\[-2pt]
\textit{A1:} ``Absolutely not. Fall’s the only season worth a damn.''\\[-2pt]
\textit{A2:} ``Absolutely. Spring’s a welcome relief after that dreary Midwest winter.''

\medskip
\textbf{\textcolor{gray!70}{\rule[0.5ex]{2em}{0.5pt}\enskip High-openness max-gap pair with high agreement (Beaches vs. Mountains, C=1)\enskip\rule[0.5ex]{2em}{0.5pt}}}
\vspace{1pt}

\noindent\textbf{A1.} man; fifties; a suburban; Southern US; Some High School; topic bias. \stg{clrState}{$P_1{=}1$} \stg{clrState}{$O_1{=}9$}.\quad \textbf{A2.} man; thirties; an exurban; Western US; Some High School; no added topic bias. \stg{clrState}{$P_2{=}5$} \stg{clrState}{$O_2{=}9$}.\\
\stg{clrOutcome}{[6] Agreement} trajectory: $A=(4,4,4,5)$.\\[-2pt]

\textit{A1:} ``I'd rather be in the mountains, to be honest. There's somethin' about the crisp air that just feels right to me.''\\[-2pt]
\textit{A2:} ``I like 'em both, but the mountains are where I'm from.''\\[-2pt]
\textit{A1:} ``I'm more of a mountain man myself.''\\[-2pt]
\textit{A2:} ``I like 'em both, but if I'm bein' honest, the mountains are more my thing.''

\end{tcolorbox}

\end{document}